\pdfoutput=1

\documentclass[11pt]{article}

\usepackage[]{acl}

\usepackage{times}
\usepackage{latexsym}

\usepackage[T1]{fontenc}

\usepackage[utf8]{inputenc}

\usepackage{microtype}

\usepackage{graphicx}
\usepackage{booktabs}
\usepackage{multirow}
\usepackage{adjustbox}
\usepackage{amsmath}
\usepackage{amssymb}

\title{On the Limitations of Reference-Free Evaluations of Generated Text}

\author{Daniel Deutsch,$^{*\dagger}$ Rotem Dror,$^\textrm{\textdaggerdbl}$ and Dan Roth$^\textrm{\textdaggerdbl}$ \\
  $^\dagger$Google Research \\
  $^\textrm{\textdaggerdbl}$University of Pennsylvania \\
  \texttt{dandeutsch@google.com} \\
  \texttt{\{rtmdrr,danroth\}@seas.upenn.edu}}

\newif\ifcomments
\commentstrue
\ifcomments
    \providecommand\dd[1]{\textcolor{blue}{[DD: #1]}}
    \providecommand\dr[1]{\textcolor{blue}{[DR: #1]}}
    \providecommand\rd[1]{\textcolor{purple}{[RD: #1]}}
    \providecommand\todo[1]{\textcolor{red}{[TODO: #1]}}
\else
    \providecommand{\dd}[1]{}
    \providecommand{\dr}[1]{}
    \providecommand{\rd}[1]{}
    \providecommand{\todo}[1]{}
\fi
  
\newcommand{\cometqe}{COMET-QE}
\newcommand{\bertscore}{BERT\-Score}
\newcommand{\qaeval}{QA\-Eval}
\newcommand{\questeval}{Quest\-Eval}
\newcommand{\summeval}{Summ\-Eval}
\newcommand{\realsumm}{REAL\-Summ}

\newcommand{\bfx}{\mathbf{x}}
\newcommand{\bfy}{\mathbf{y}}
\newcommand{\hatbfy}{\hat{\mathbf{y}}}
\newcommand{\starbfy}{\mathbf{y}^*}
\newcommand{\refbased}{\mathcal{M}_\textrm{Ref-Based}}
\newcommand{\reffree}{\mathcal{M}_\textrm{Ref-Free}}

\newcommand\blfootnote[1]{%
  \begingroup
  \renewcommand\thefootnote{}\footnote{#1}%
  \addtocounter{footnote}{-1}%
  \endgroup
}

\begin{document}
\maketitle

\begin{abstract}
    There is significant interest in developing evaluation metrics which accurately estimate the quality of generated text without the aid of a human-written reference text, which can be time consuming and expensive to collect or entirely unavailable in online applications.
    However, in this work, we demonstrate that these reference-free metrics are inherently biased and limited in their ability to evaluate generated text, and we argue that they should not be used to measure progress on tasks like machine translation or summarization.
    We show how reference-free metrics are equivalent to using one generation model to evaluate another, which has several limitations:
    (1) the metrics can be optimized at test time to find the approximate best-possible output,
    (2) they are inherently biased toward models which are more similar to their own,
    and (3) they can be biased against higher-quality outputs, including those written by humans.
    Therefore, we recommend that reference-free metrics should be used as diagnostic tools for analyzing and understanding model behavior instead of measures of how well models perform a task, in which the goal is to achieve as high of a score as possible.\footnote{\url{https://cogcomp.seas.upenn.edu/page/publication_view/991}}  
\end{abstract}
\blfootnote{$^*$Work done while at the University of Pennsylvania}
\section{Introduction}

Automatically evaluating the quality of generated texts is essential for the development of natural language generation systems.
The most common type of evaluation for generation tasks such as machine translation (MT) and summarization is done with reference-based automatic metrics, which evaluate a text by comparing it to a gold-standard reference text, usually written by humans \citep[][\emph{inter alia}]{PRWZ02,Lin04,ZKWWA20,SellamDaPa20,DeutschBeRo21,ZhangBa21}.

Reference texts can be expensive to collect or are entirely unavailable when there is a need to estimate the quality of text in real time, so there is an increased interest in developing automatic metrics that do not use references to evaluate text, commonly referred to as reference-free metrics \citep[][\emph{inter alia}]{LouisNe13,FYMFF19,SLPS19,SDLPSWG21,VasilyevDhBo20,RFZSSRGML21}.
While these metrics do not always achieve performance parity with their reference-based counterparts, their high correlations to human judgments suggest that reference-free evaluation is a promising direction of future research \citep{FYMFF19}.\footnote{Some reference-free metrics actually already out-perform reference-based metrics \citep{FRMLSFLB21}.}

\begin{figure}
    \centering
    \begin{adjustbox}{width=\columnwidth}
    \begin{tabular}{llc}
        \toprule
        & & \bf Prism-src ($\uparrow$) \\
        \midrule
        \multirow{2}{*}{\bf Source} & \multirow{2}{5cm}{Doch er ist nicht krank, er hat nur einen m\"{a}chtigen \underline{\color{red} Kater}.} &  \\
        \\
        \multirow{2}{*}{\bf Reference} & \multirow{2}{5cm}{But he is not ill, he only has quite a \underline{\color{red} hangover}.} & \multirow{2}{*}{-1.6} \\
        \\
        \multirow{2}{*}{\bf Candidate} & \multirow{2}{5cm}{But he is not sick, he has only one powerful \underline{\color{red} cat}.}  & \multirow{2}{*}{-0.4} \\
        \\
        \midrule
        \multirow{2}{*}{\bf Source} & \multirow{2}{5cm}{Und \underline{\color{red}mit Mann und Maus} gegen Mainz verteidigt.} &  \\
        \\
        \multirow{2}{*}{\bf Reference} & \multirow{2}{5cm}{And \underline{\color{red}threw everything they had} into our defense.} & \multirow{2}{*}{-4.8} \\
        \\
        \multirow{2}{*}{\bf Candidate} & \multirow{2}{5cm}{And defended \underline{\color{red}with man and} \underline{\color{red}mouse} against Mainz.} & \multirow{2}{*}{-0.4} \\
        \\
        \bottomrule
    \end{tabular}
    \end{adjustbox}
    \caption{
        Here, Prism-src was optimized to generate the candidate translations.
        They are clearly wrong (\emph{Kater} means both ``cat'' and ``hangover'';
        \emph{mit Mann und Maus} is an expression that means ``with all means available''), but have better Prism-src scores than the references.
        Comparing systems with reference-free metrics will favor systems that are more similar to the metrics' underlying models rather than higher quality output.
    }
    \label{fig:examples}
\end{figure}

However, in this work, we demonstrate that reference-free evaluation metrics have inherent limitations and argue that they should not be used to measure progress on tasks, even in domains in which no reference texts are available.
Central to our argument is the idea that because reference-free metrics evaluate text using the same input provided to the generation models, they are either explicitly or implicitly using an underlying generation model to evaluate other models (\S\ref{sec:ref_free}).
There are several implications of this, which we explore through an analysis of three reference-free evaluation metrics, Prism-src \citep{ThompsonPo20} and \cometqe{} \citep{RFZSSRGML21} for MT and \questeval{} \citep{SDLPSWG21} for summarization.

First, the metrics' underlying models will achieve the best possible metric score by definition.
Therefore, the ``perfect'' model is already known, and we show that it is possible to define simple approximate inference algorithms which use these models to find the approximate best output according to the metrics (\S\ref{sec:metric_opt}, \S\ref{sec:inference_eval}).

Then, the metrics have inherent, undesirable biases that originate from their underlying models.
Not only do they favor the underlying models' outputs, but they are also biased toward outputs from models which are similar to their own, and biased against higher-quality outputs, such as those written by humans (Fig.~\ref{fig:examples}, \S\ref{sec:biases}, \S\ref{sec:pseudo_ref}).
Thus, if they were used as primary evaluation methods for a task, they would encourage other models to be more similar to their own and less human-like, an undesirable property of an evaluation metric.

Our recommendation is that reference-free metrics should not be used as methods for measuring progress on generation tasks such as MT, in which the goal is to achieve the highest possible value of the metric.
Instead, they are better suited to be diagnostic statistics for analyzing model behavior with the understanding that they are inherently limited and biased (\S\ref{sec:discussion}).

The contributions of this work include:
(1) insight on the equivalence of reference-free metrics and generation models,
(2) a demonstration that reference-free metrics' values can be optimized at test time to achieve high-scoring outputs, and (3) an analysis that reveals reference free metrics' inherent biases and limitations.

\section{Reference-Free Metrics as Models}
\label{sec:ref_free}
Conditional text generation models can be viewed as a function $\theta(\cdot)$ which scores an output text $\bfy \in \mathcal{Y}$ for some input text $\bfx$.
Then $\theta(\cdot)$ is used in conjunction with an inference procedure $f_\theta(\cdot)$ to find the best output at test time.\footnote{
    In practice, $f_\theta(\cdot)$ finds the 
    \emph{approximate} best output, not the global maximum of $\theta(\cdot)$.
}
\begin{align}
    \theta(\bfx, \bfy) &\rightarrow \mathbb{R} \\
    f_\theta(\bfx) &= \underset{\bfy \in \mathcal{Y}}{\arg\max{}} \; \theta(\bfx, \bfy)
\end{align}
For instance, $\theta(\cdot)$ could be a learned sequence-to-sequence model and $f_\theta(\cdot)$ could be beam search.

The output of $f_\theta(\cdot)$, denoted $\hatbfy$, is typically evaluated by some automatic metric $\mathcal{M}$.
Reference-based metrics do this by scoring $\hatbfy$ using some gold-standard text $\starbfy$ (which is not available to the model during inference) and the input $\bfx$ (which is not always used).
For instance, $\refbased$ could calculate a BLEU score \citep{PRWZ02} between the output translation $\hatbfy$ and the gold translation $\starbfy$.
\begin{equation}
    \refbased(\bfx, \hatbfy, \starbfy) \rightarrow \mathbb{R}
\end{equation}
In contrast, reference-free metrics calculate a score for $\hatbfy$ without $\starbfy$:
\begin{equation}
    \reffree(\bfx, \hatbfy) \rightarrow \mathbb{R}
\end{equation}
Such metrics include the three analyzed in this work, namely, Prism-src \citep{ThompsonPo20}, \cometqe{} \citep{RFZSSRGML21}, and \questeval{} \citep{SDLPSWG21}.

Because $\theta(\cdot)$ and $\reffree$ are both functions of only $\bfx$ and $\bfy$ (equivalently $\hatbfy$), $\reffree$ itself can be viewed as a conditional generation model.
For some metrics, such as Prism-src, this is explicitly stated, whereas others are implicitly making this assumption.
This is not the case for reference-based metrics since they additionally require $\starbfy$ as input.

Since reference-free metrics are equivalent to generation models, there must exist some inference procedure which finds the best output text under the metric, denoted $g_{\reffree}(\cdot)$:
\begin{equation}
    g_{\reffree}(\bfx) = \underset{\bfy \in \mathcal{Y}}{\arg\max} \; \mathcal{M}_\textrm{Ref-Free}(\bfx, \bfy)
\end{equation}
Computing $g_{\reffree}(\cdot)$ may be computationally expensive because $\reffree$ may not support efficient inference.
However, the inference procedure does always exist, and will return the best possible output according to the reference-free metric by definition.

We explore the implications of using a model to evaluate other models by analyzing the behavior of three different reference-free evaluation metrics on two text generation tasks, MT and summarization.

\section{Analysis Setup}
\label{sec:setup}

Here, we discuss the datasets and metrics used in our analysis of reference-free metrics.

\paragraph{Datasets}
Our MT experiments are run on the data collected for the WMT'19 metrics shared task \citep{MWBG19}, which includes reference translations and human-judged model outputs for 10 to 20 translation systems across 18 language pairs.

The summarization experiments use the \summeval{} \citep{FKMSR21} and \realsumm{} \citep{BGALN20} datasets, which consist of reference summaries and human-judged model outputs for 16 and 25 summarization models, respectively, collected from the CNN/DailyMail dataset \citep{NZSGX16}.

\paragraph{Prism-src}
Prism-src is a reference-free evaluation translation metric that scores a translated text according to the log-probability of the translation conditioned on the original source text under a learned sequence-to-sequence translation model \citep{ThompsonPo20}.
The model is a multi-lingual MT model, meaning it was trained using many different language pairs, so the same learned parameters can be used to score translations in various languages.

\paragraph{\cometqe{}}
\cometqe{} \citep{RFZSSRGML21} is a modification of the learned reference-based MT evaluation metric COMET \citep{RSFL20}.
COMET embeds the candidate translation, source text, and reference translation using a cross-lingual encoder, creates a pooled featured representation using the three encodings, and trains the model end-to-end to predict human judgments of the quality of the candidate translation.
\cometqe{} uses the same architecture to predict a score for the candidate translation but only uses the candidate translation and source text to create the pooled feature representation, and is therefore reference-free.

\paragraph{\questeval{}}
\citet{SDLPSWG21} proposed a reference-free summarization metric called \questeval{} which generates QA pairs from both the source document and generated summary then scores the summary based on the proportion of those pairs which are answered correctly in the opposite text.
The metric optionally includes a step in which the QA pairs generated from the source document are weighted based on a learned query weighting model.
The query weighter was trained to predict the probability that a question is answered in the CNN/DailyMail reference summaries using a pre-trained QA model.
We use the query weighter in our experiments since it improved the performance of \questeval{} in \citet{SDLPSWG21}.

\paragraph{Reference-Based Metrics}
We analyze the reference-free metrics with respect to various reference-based metrics which have been demonstrated to have strong correlations to human judgments of translation/summary quality.
BLEU \citep{PRWZ02} and ROUGE \citep{Lin04} compare the two texts using $n$-gram overlap statistics.
\bertscore{} calculates a quality score based on how similar the reference and candidate texts' BERT \citep{DCLT19} embeddings are \citep{ZKWWA20}.
\qaeval{} is a QA-based metric for summarization, which generates wh-questions from the reference summary and calculates a score for the candidate summary based on the proportion of questions answered correctly \citep{DeutschBeRo21}.
Finally BLEURT is a learned MT metric which predicts a translation quality score using encoded BERT representations of the reference and candidate translations \citep{SellamDaPa20}. \\

Implementation details can be found in Appendix~\ref{sec:implementation_details}.
\section{Metric Optimization}
\label{sec:metric_opt}

Since reference-free metrics are equivalent to models, then it is possible to define inference procedures which produce the best-possible outputs according to the metrics.
Here, we discuss three such (approximate) inference procedures.
Importantly, they can all be run at test time because they do not rely on a reference text.

\subsection{Direct Optimization}
\label{sec:direct_opt}
If a reference-free metric scores a candidate output in a way that an efficient approximate inference procedure can be defined, then finding the best possible output under the metric is straightforward.

Among the metrics analyzed in this paper, only Prism-src falls into this category.
Because Prism-src assigns a score to a translation equal to its average log-probability under a learned sequence-to-sequence MT model, the approximate best translation under Prism-src can be found by running beam search with the MT model conditioned on the source text.

\subsection{Greedy Optimization for Extractive Summarization}
\label{sec:greedy_ext}
Summarization models are generally categorized as being either extractive or abstractive.
Extractive systems create a summary by selecting $k$ salient document sentences, whereas abstractive systems typically autoregressively generate a summary with a sequence-to-sequence model.

The best possible extractive summary according to a reference-free metric can be found by enumerating all possible summaries of $k$ sentences, scoring them with the metric, and selecting the summary with the highest score.
Since the number of $k$ sentence summaries may be large, this may be computationally expensive.
However, an approximate inference procedure can be used instead.

Rather than enumerate all possible extractive summaries, the approximate inference algorithm constructs a summary by greedily selecting one sentence that increases the score of the metric the most \citep{LinBi11}.
This is repeated until a target summary length of $k$ sentences is reached, resulting in an approximation of the best possible summary under the reference-free metric.

A near-identical procedure is commonly used for creating sentence-level labels for training extractive summarization models, except a reference-based evaluation metric, such as ROUGE, is typically used for scoring the sentences instead of a reference-free metric \citep{NallapatiZhZh17}.
The key difference is that the output summary from the reference-based procedure is used to train a model which later predicts $k$ salient sentences during inference, whereas the reference-free procedure can be directly used during inference (i.e., without training) to pick the approximately best summary under the reference-free metric.

\subsection{Reranking}
\label{sec:reranking}
Exact inference for any reference-free metric can be performed by enumerating all possible outputs, calculating the score of each one, and selecting the output with the highest score.
However, it is almost certainly true that this is computationally intractable for any practical application of text generation due to the size of the output space.

To that end, we propose to use reranking \citep{ShenSaOc04,OGKSYFKSSEJJR04} as an approximate inference procedure in which a pre-trained model for the task at hand is used to restrict the search space to a small set of high-quality candidate outputs.
These outputs are then scored and reranked using the reference-free metric to identify an approximately best output under the metric.

In practice, we identify a set of $k$ high-quality outputs using standard beam search with pre-trained sequence-to-sequence summarization and MT models and a beam size of $k$.
The top-$k$ partial outputs sorted by their log-likelihood under the pre-trained models are kept at each step of beam search.
The final outputs are then reranked by a reference-free metric.
For summarization, we use BART \citep{LLGGMLSZ20} trained on the CNN/DailyMail dataset.
For MT, we use Facebook's submission to the WMT'19 translation shared task \citep{NYBOAE19}.
The model is available for en$\rightarrow$de, de$\rightarrow$en, en$\rightarrow$ru, and ru$\rightarrow$en.

\section{Analysis}
\label{sec:results}

\subsection{Approximate Inference Effectiveness}
\label{sec:inference_eval}
Although inference methods for the reference-free metrics can be defined, it is possible that they fail to find high-scoring outputs due to the complexity of the search problem.
However in this analysis, we show that the simple approximate inference procedures defined in \S\ref{sec:metric_opt} are effective at optimizing the metrics' scores.

\begin{figure*}[t]
    \centering
    \includegraphics[width=\textwidth]{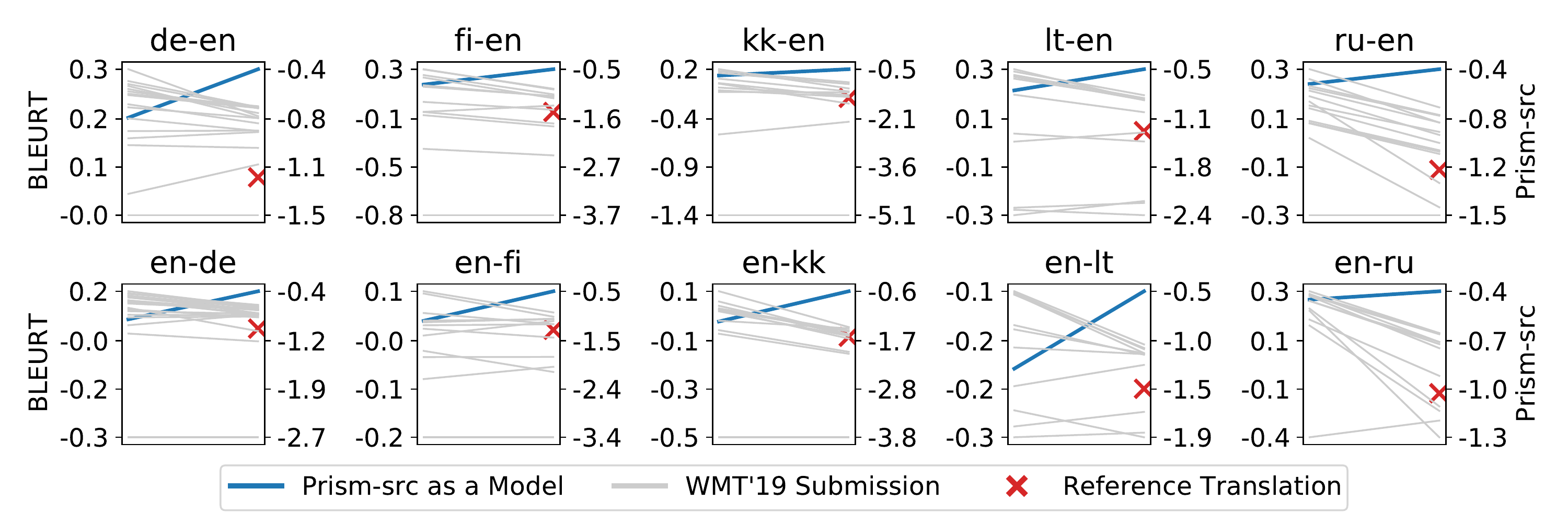}
    \caption{
        Each line in these plots corresponds to one system evaluated under two different metrics on either y-axis.
        This illustrates the change in system ranking between the two metrics.
        We see that directly optimizing Prism-src (blue line; \S\ref{sec:direct_opt}) yields the highest Prism-src performance (right y-axis) but only an average system as evaluated by BLEURT (left y-axis).
        The reference translation (red ``x'') has a lower Prism-src score compared to many systems across all language pairs, demonstrating Prism-src's biases toward learned model output and against human-written translations.
    }
    \label{fig:ranking_prism_bleurt_subset}
\end{figure*}

\begin{figure}[t]
    \centering
    \includegraphics[width=\columnwidth]{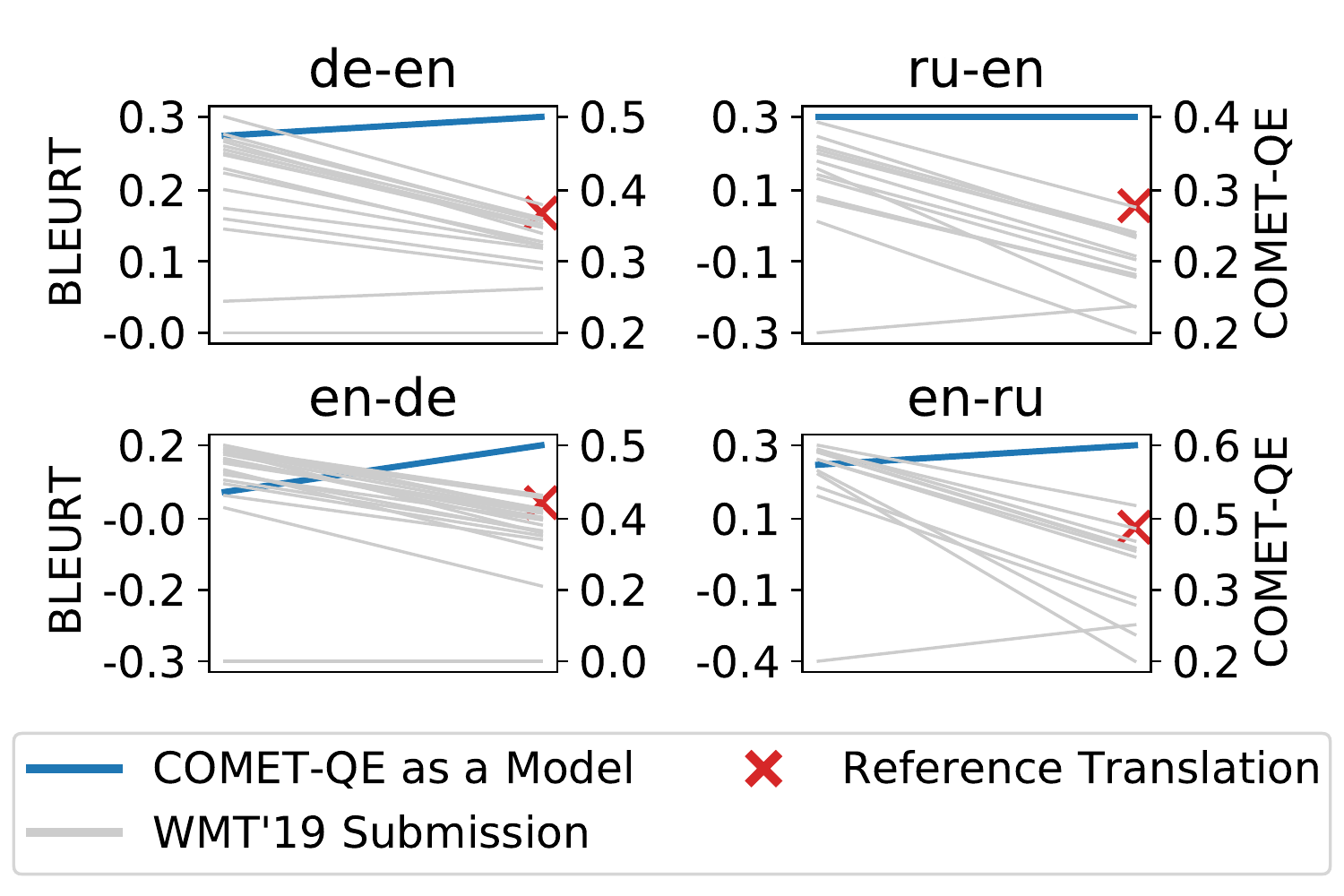}
    \caption{
    Reranking the output of a pre-trained model results in COMET-src scores which are far higher than the reference translations' scores (red ``x''), demonstrating that a better COMET-src value means the translation is more similar to the metric's underlying model instead of human-written text.
    }
    \label{fig:ranking_comet_bleurt}
\end{figure}

We compared the outputs obtained by the inference algorithms to those from systems included in the WMT'19, SummEval, and REALSumm datasets.
Fig.~\ref{fig:ranking_prism_bleurt_subset} evaluates using the direct optimization procedure (\S\ref{sec:direct_opt}) to select the best Prism-src output, Fig.~\ref{fig:ranking_comet_bleurt} shows the results of using reranking (\S\ref{sec:reranking}) to pick the best outputs according to \cometqe{}, and Fig.~\ref{fig:ranking_questeval_rouge} contains the results of using the greedy extractive procedure (\S\ref{sec:greedy_ext}) to optimize QuestEval.
The Figures also include the systems' scores under the reference-based metrics BLEURT for MT and ROUGE for summarization.
Other combinations of reference-based metrics and inference algorithms can be found in Appendix~\ref{sec:additional_results}.


In all MT language pairs and both summarization datasets, the inference algorithms produce the highest scoring outputs under the reference-free metrics, often by a large margin.
For example, reranking translations according to their \cometqe{} scores on de$\rightarrow$en results in a relative 38\% improvement in \cometqe{} over the WMT'19 submission with the best \cometqe{} score (0.347 $\rightarrow$ 0.478).
Clearly, the simple inference procedures can be used to find very high scoring outputs under the reference-free metrics even if the metric does not directly support efficient inference.

Despite the improvements in reference-free scores, it does not appear as if these outputs are as high-quality as the reference-free metric scores would indicate.
Ideally, the outputs would be rated by humans to establish a ground-truth quality score, but a fair comparison to the other systems' outputs included in the datasets would require re-judging their translations or summaries, which is prohibitively expensive.
Instead, we use reference-based metrics as indicators of quality.

When the outputs from our inference algorithms are compared to other systems using reference-based metrics (also shown in Figs.~\ref{fig:ranking_prism_bleurt_subset}, \ref{fig:ranking_comet_bleurt}, and \ref{fig:ranking_questeval_rouge}), we see that they are often of average quality or worse.
For example, the greedy extractive summaries obtained by optimizing \questeval{} on \realsumm{} are among the lowest-performing in the dataset according to ROUGE-2 (Fig.~\ref{fig:ranking_questeval_rouge}).
Thus, directly optimizing the reference-free metrics does not always yield a high-quality system, at least according to reference-based metrics (examples shown in Fig~\ref{fig:examples}).
 
\subsection{Undesirable Metric Biases}
\label{sec:biases}

\begin{figure}[t]
    \centering
    \includegraphics[width=\columnwidth]{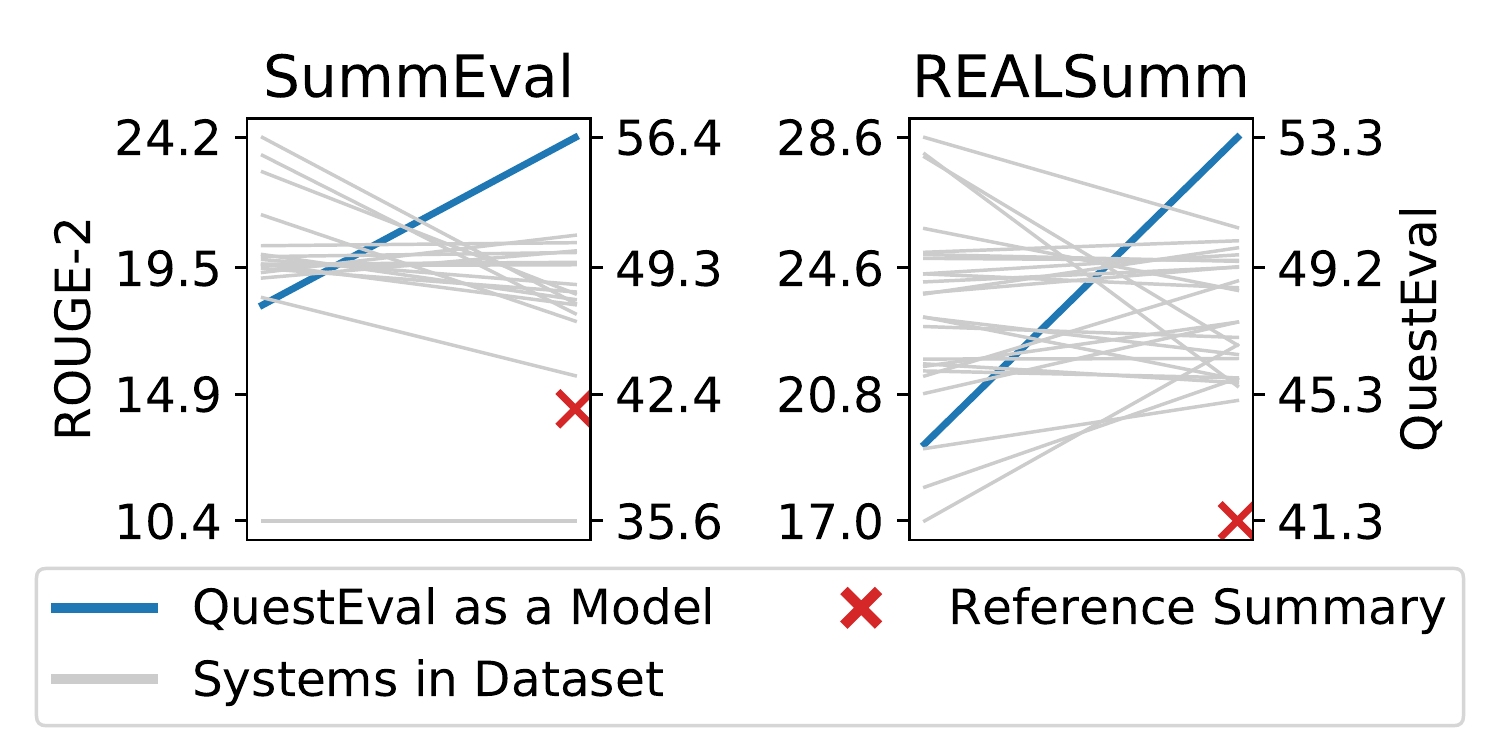}
    \caption{
        Nearly all models in the Summ\-Eval and REAL\-Summ datasets have better QuestEval scores than the reference summaries, demonstrating the metrics' bias toward learned model output over human-written text.
    }
    \label{fig:ranking_questeval_rouge}
\end{figure}

Ideally, evaluation metrics should score human-written text higher than learned model outputs since it is very likely that the human references are of higher-quality.
However, we see that this does not always happen with reference-free metrics.

Figs.~\ref{fig:ranking_prism_bleurt_subset}, \ref{fig:ranking_comet_bleurt}, and \ref{fig:ranking_questeval_rouge} additionally contain the scores of the reference texts under the reference-free metrics (marked with a red ``x'').
In all settings, the inference algorithms' outputs score higher than the references.
This is unsurprising because they select their outputs to optimize the metrics' values.
However, it demonstrates that as models continually improve their reference-free scores, they will begin to converge onto the metrics' underlying models and not onto human-quality text.
As the research goal of text generation is to build models which produce human-like text, the reference-free metrics' scores do not align with this goal.

Further, the same figures show that it is often the case that other models in the datasets---which did not directly optimize the reference-free metrics---also are scored higher than human text by the reference-free metrics.
This is especially true for Prism-src and QuestEval, but less so for \cometqe{}.
For example, on language pair de-en, only one system has a lower Prism-src score than the reference translations (Fig.~\ref{fig:ranking_prism_bleurt_subset}).
These metrics appear to have a bias for outputs from learned models over human-written text.

\begin{figure*}
    \centering
    \includegraphics[width=\textwidth]{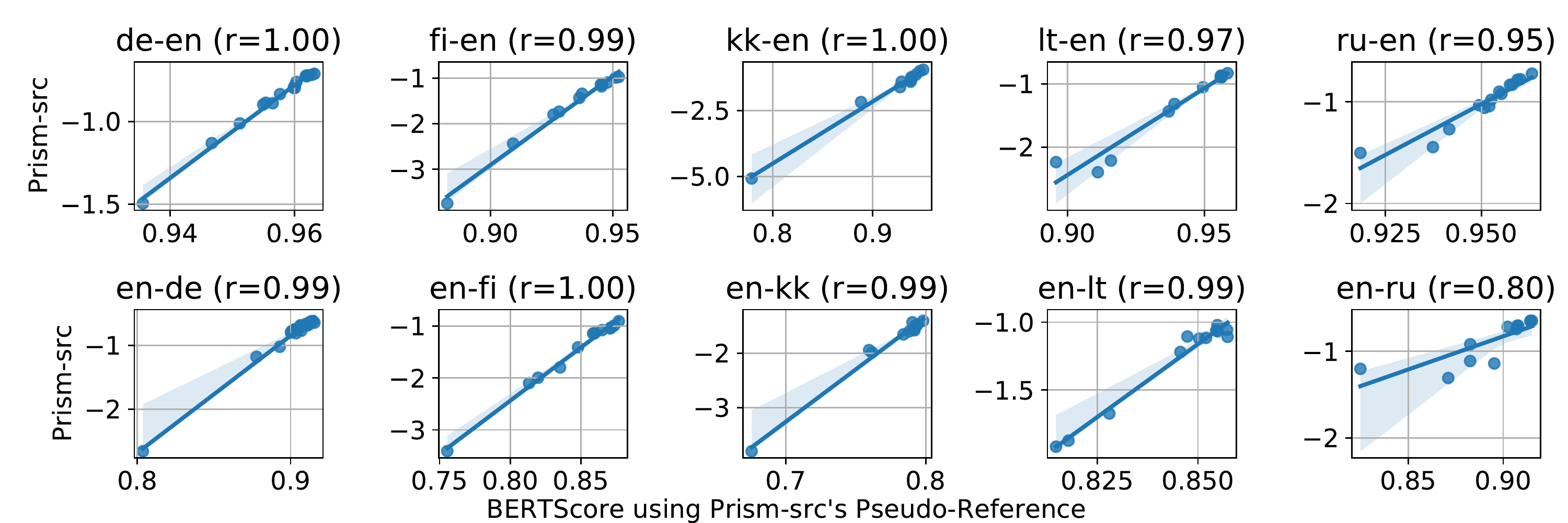}
    \caption{
        A system's \bertscore{} in which the output from directly optimizing Prism-src is used as the reference (x-axis) is strongly correlated to that same system's Prism-src score (y-axis).
        This demonstrates Prism-src is roughly equivalent to evaluating systems with a pseudo-reference translation which is generated by a model.
        Pearson's $r$ shown in the title of each plot.
    }
    \label{fig:bertscore_prism_xbleu}
\end{figure*}
\begin{figure}
    \centering
    \includegraphics[width=\columnwidth]{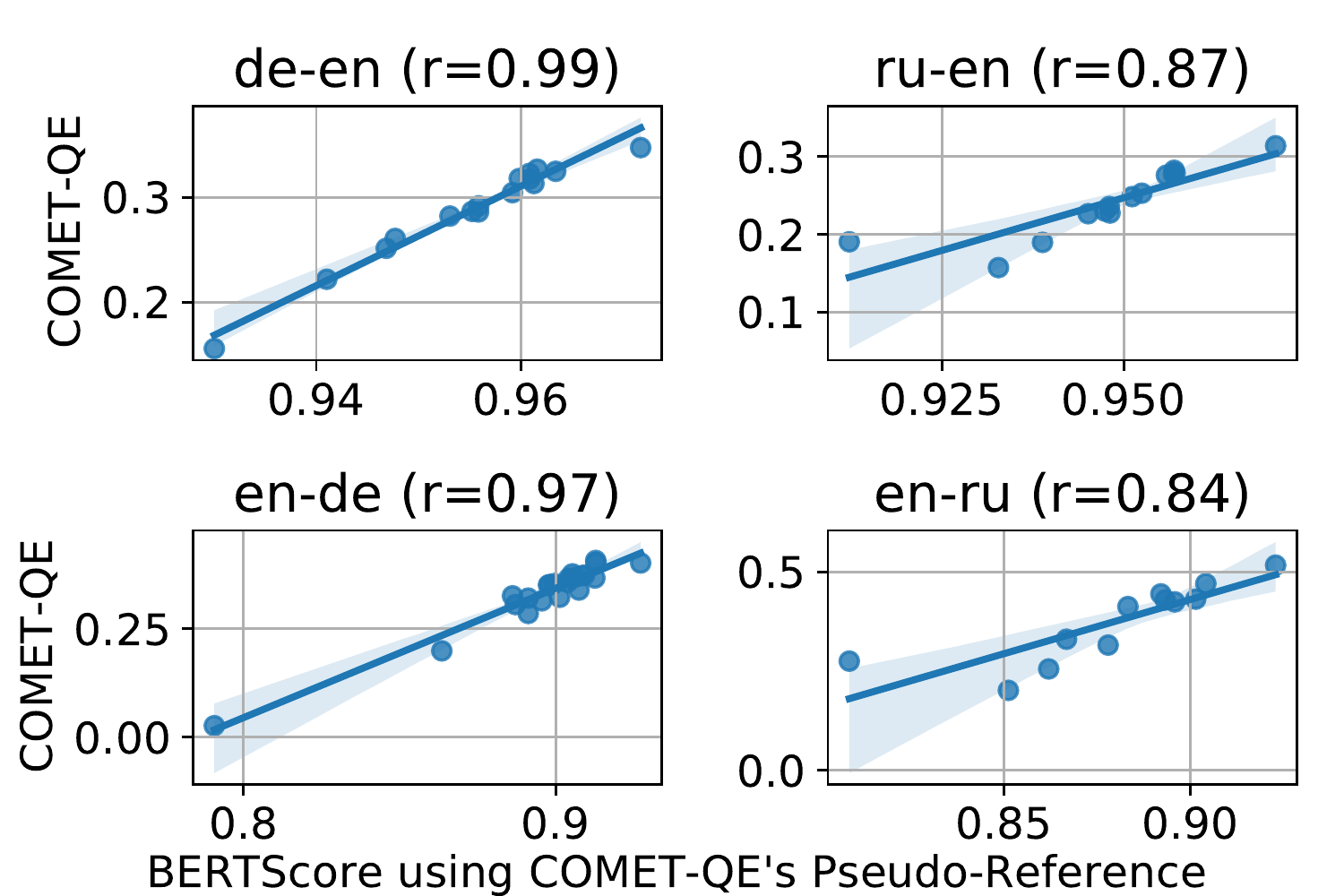}
    \caption{
        Scoring systems with \bertscore{} against psudeo-references obtained by optimizing COMET-src strongly correlates to the systems' COMET-src scores.
    }
    \label{fig:bertscore_comet_xbleu}
\end{figure}

We suspect this bias for outputs from learned models is due to how the internal models used by Prism-src and QuestEval to score text are trained.
Prism-src scores translations using an MT model that was trained on standard MT data.
Quest\-Eval uses a question-weighting model that predicts how likely a question is answered in a reference summary, which was trained on the same CNN/DailyMail dataset that the summarization models in the summarization datasets were also trained on.
Thus, these metrics internally use models which are directly or indirectly trained to perform the generation task (MT or summarization).

Generation models which are trained on the same datasets are known to exhibit similar behavior and make similar mistakes, and their outputs often look markedly different from human-written text.
Therefore, we suspect that the signals the internal MT/question-weighting models have learned to identify high-quality text are similar to those which the task-specific models have learned to produce their output, and thus receive high scores by the metrics.
In contrast, the human-written text likely does not rely on these signals, and is thus perceived as low-quality by the metrics.

This bias toward learned model outputs is potentially less severe for \cometqe{} because unlike Prism-src and Quest\-Eval, it is specifically trained to predict translation quality using manually collected human judgments.
Those human judgments also contain evaluations of human-written translations, so it is better at distinguishing human-versus-model output \citep{FRMLSFLB21}.
Thus, the signals it learns to identify high-quality text are likely different than what is learned by the translation models in the WMT'19 dataset.

In summary, the metrics' biases toward their own outputs (or other learned model outputs) and against human texts demonstrates they reward outputs which look more like their own instead of human-quality text, an undesirable property of an evaluation metric.

\subsection{Reference-Free Metrics as Pseudo-References}
\label{sec:pseudo_ref}
Thus far, we have argued that the underlying model of a reference-free metric is the theoretical best model according to the metric.
It would intuitively follow that the more similar another model is to the metric's underlying model, the higher metric score that model would receive.
To that end, in this analysis we demonstrate that scoring a system with a reference-free metric is roughly equivalent to evaluating that system's outputs against the metric's best possible output using a reference-based metric.
This further demonstrates the limitations of reference-free metrics, including their biases toward their own underlying models' outputs.

A pseudo-reference is a piece of text which is used in place of a human-written reference to evaluate some candidate text with a reference-based metric \citep{LouisNe13,GaoZhEg20}.
For the reference-free metrics, we define the pseudo-reference to be the output from the inference procedures defined in \S\ref{sec:metric_opt} (i.e., those evaluated in \S\ref{sec:inference_eval}).
For example, the Quest\-Eval pseudo-reference is the extractive summary which was selected to greedily maximize the Quest\-Eval score.

\begin{figure}
    \centering
    \includegraphics[width=\columnwidth]{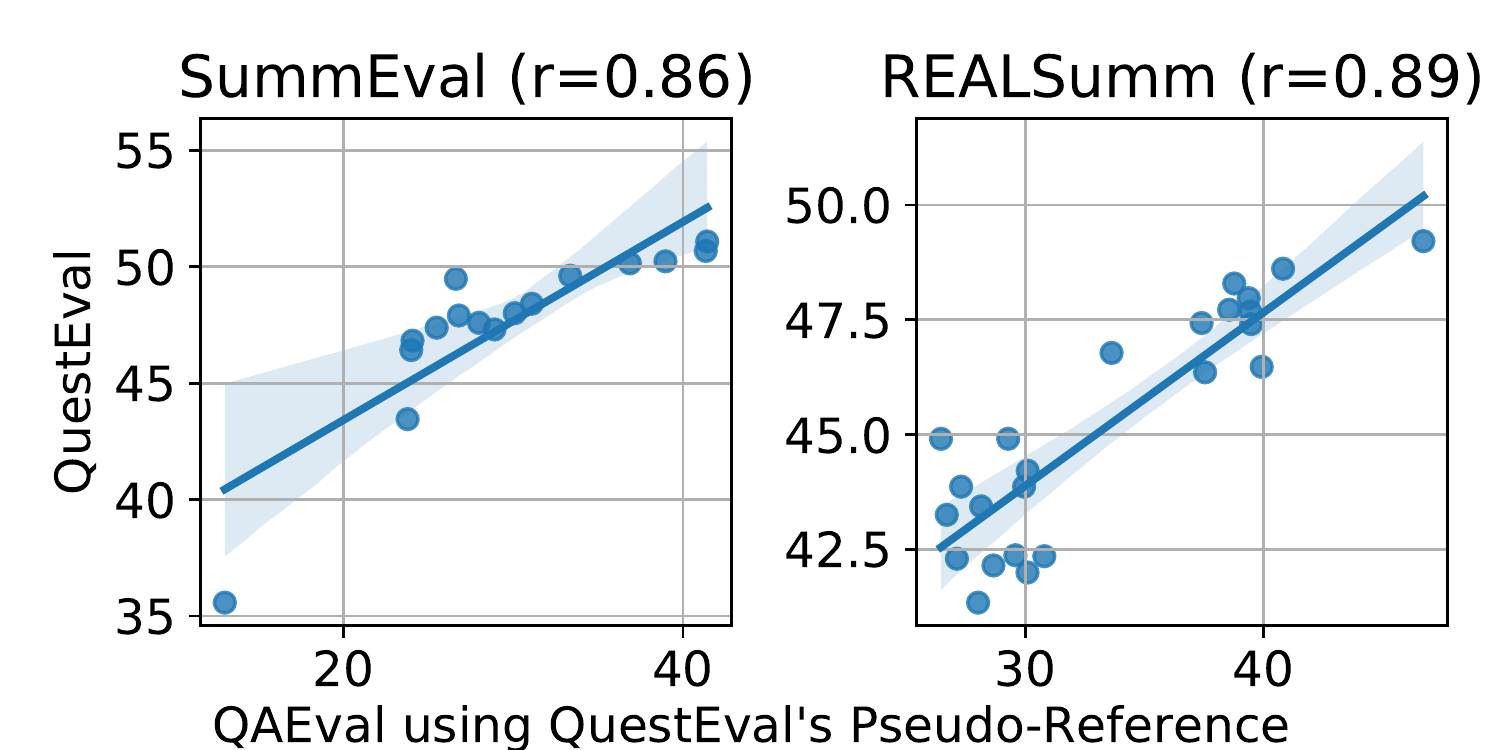}
    \caption{
        Calculating a system's \qaeval{} score against the psuedo-reference chosen to maximize its Quest\-Eval score is strongly correlated with that same systems' Quest\-Eval score on Summ\-Eval and REAL\-Summ.
    }
    \label{fig:xrouge_qaeval_questeval}
\end{figure}

Once the pseudo-references have been defined, they can be used in conjunction with any reference-based metric, such as BLEURT or \qaeval{}, to evaluate other translations or summaries.
To quantify the similarity between evaluating a system with a pseudo-reference and a reference-free metric, we calculated the Pearson correlation between the system-level scores between the two methods.
These correlations are show in Figs.~\ref{fig:bertscore_prism_xbleu}, \ref{fig:bertscore_comet_xbleu}, and \ref{fig:xrouge_qaeval_questeval}.

For MT, we find that the \bertscore{} of a translation that uses pseudo-references instead of the human-written reference has an average Pearson correlation of 0.95 and 0.92 to the Prism-src (Fig.~\ref{fig:bertscore_prism_xbleu}) and \cometqe{} scores (Fig.~\ref{fig:bertscore_comet_xbleu}), respectively.
The summarization correlations for Quest\-Eval to \qaeval{} using a pseudo-reference (Fig.~\ref{fig:xrouge_qaeval_questeval}) are also rather strong at 0.88 on average.
The correlations using other reference-based metrics are slightly weaker on average due to low values on specific datasets or language pairs (see Appendix~\ref{sec:additional_results}), but there are many instances in which the correlations are $\geq 0.9$.

Overall, these correlations are very strong, suggesting that the reference-free metrics are roughly equivalent to using pseudo-references to evaluate other models.
Once the metrics are viewed this way, their limitations become clear.
The metrics' outputs are the gold-standard against which all other outputs should be evaluated.
Thus, the metrics favor their own outputs (the pseudo-references) and other outputs which are more similar (where similarity is measured using reference-based metrics).
Further, their ability to evaluate other models is inherently limited by the qualities of their pseudo-references.
If a system outputs a translation or a summary which is higher-quality than the pseudo-reference, it will be incorrectly penalized because it is different than the pseudo-reference even though those differences are actually improvements.
Thus, the metrics' scores of systems which are better in quality than their own underlying models will be misleading.

\section{Discussion}
\label{sec:discussion}

\subsection{Reference-Free Evaluation}

Due to the limitations of reference-free metrics that we demonstrated in previous Sections, we argue that reference-free metrics should not be used to measure progress on a task, for instance, by concluding that an MT or summarization model is better than another because it has a higher reference-free metric score.
If they are used this way, the model which will achieve the best performance is already known (the metric's underlying model; \S\ref{sec:ref_free}), and simple inference procedures are effective at using those models to generate high-scoring outputs (\S\ref{sec:metric_opt}, \S\ref{sec:inference_eval}).
Improving the metric's score is no longer about creating better models on the training data for the task;
rather, it is about coming up with better procedures to optimize the metric during inference.
In the end, the quality of the system which is produced by this evaluation methodology is limited by the quality of the metric itself.

When references are not available, we recommend investing resources into collecting human-written references that can be used for evaluation purposes instead of pursuing reference-free metrics.
Although individual reference-based metrics have their own flaws \cite{Graham15, BGALN20, DeutschDrRo21}, the class of reference-based metrics still ultimately encourages systems to generate text which is similar to humans, which is the goal of text generation research as it is currently defined.

\subsection{What about Inherently Reference-Free Evaluations?}
Although we argue that measuring system quality with reference-free metrics is flawed and misleading, abandoning reference-free evaluations completely is not an option and one which we do not advocate for.
There are many aspects of generated text that inherently do not rely on the presence of a reference to be evaluated, such as the fluency of a translation or how faithful a summary is to its input document.
There is no obvious benefit of including a reference text in these evaluations.
As such, they suffer from the same issues as the metrics analyzed in this work, yet the motivation for being able to automatically evaluate these aspects of text is clear.

In these scenarios, we recommend using reference-free metrics as diagnostic tools for better understanding the behavior of models instead of a method for measuring progress on a task.
For instance, the perplexity of a summary under a large-scale language model is a useful statistic to report in order to approximate its fluency, but the value should only be interpreted as exactly what it measures---how likely the observed text is under the language model---with the understanding that the measure is inherently biased towards the underlying language model.
The perplexity should not be used to drive research on how to generate more fluent summaries because the most fluent summarization model is the language model itself.

This recommendation applies not only to metrics which measure inherently reference-free aspects of text, but also to the metrics that evaluate aspects which we argue should use references, such as those analyzed in this work.
They are certainly useful statistics to report, but improving their values as much as possible should not be the goal.
\section{Related Work}
Various other reference-free metrics have been proposed for MT \citep{FYMFF19,RFZSSRGML21}, summarization \citep{LouisNe13,SLPS19,XMAA19,VasilyevDhBo20,SDLPSWG21},
dialog generation \citep{MehriEs20,HCANSA21}, image captioning \citep{HHFBC21}, and simplification \citep{MHMCBS18,KrizApCa20}.
Reference-free metrics for MT were evaluated and compared to reference-based metrics in the WMT'21 metrics shared task \citep{FRMLSFLB21}.
In some evaluation settings, reference-free metrics perform better than reference-based metrics.

While many of the work which propose reference-free metrics do not explicitly state that they are trying to replace reference-based metrics, we worry that this goal is implied and that the limitations of their metrics are not stated clearly enough.
For example, simply by arguing that their reference-free metrics have stronger correlations to human judgments than their reference-based counterparts, an uninformed reader may conclude that the two types of metrics are interchangeable and use a reference-free metric instead of a reference-based metric without fully understanding the consequences.
We recommend authors be explicit about the limitations of reference-free metrics and say, for example, they can be used to complement existing reference-based evaluations rather than a replacement \citep{LouisNe13}.

One such application of reference-free metrics which fits our recommendations for how they should be used is the WMT shared task on quality estimation \citep[QE; ][]{SBFFCGM20,SBFZLCM21}.
QE metrics are used to flag translations which might require post-editing or to identify ``catastrophic errors'' in translations, for instance hateful or violent text that was hallucinated by the model, without the aid of a reference.
The intended purpose of QE metrics is not to rank and compare MT systems.
We argue that the ``rebranding'' of QE metrics as reference-free metrics by some authors contributes to the notion that they can and should be used the same way that reference-based metrics are and that they are inherently better because they do not require a reference.
We recommend that future proposals of methods for evaluating generated text without references are called QE metrics instead of reference-free metrics to make the distinction more clear.

In previous work, Prism-src was further explored by \citet{AFFC21}, who experimented with its model capacity, scoring methods, and more.
They perform an analysis using pseudo-references similar to ours in \S\ref{sec:pseudo_ref}.
In their experiment, they show that the cross-BLEU score calculated between Prism-src's pseudo-reference and the system output does not appear to be correlated to Prism-src's system ranking relative to human-based system ranking.
That is, a higher cross-BLEU score does not necessarily result in Prism-src ranking a system higher than it should.
They conclude that their analysis does not show evidence that Prism-src is biased toward outputs which are more similar to its own, which seemingly contradicts our own experiments and results in \S\ref{sec:pseudo_ref}.
However, we argue that the high correlations between a system's Prism-src's score and its similarity to the metric's pseudo-reference (measured by reference-free metrics) does demonstrate such a bias exists, but \citet{AFFC21} did not find evidence this bias negatively impacted the rankings of the systems.

Our experiments in \S\ref{sec:reranking} leverage reranking to optimize reference-free metrics' scores.
Reranking in machine translation has a long history \citep{ShenSaOc04,OGKSYFKSSEJJR04} and has been done with reference-free metrics, for instance in concurrent work by \citet{FeFRSONM22}.
The aim of our experiment was not to propose a novel technique, but rather to point out that reference-free metrics can be optimized in this way, and doing so may be potentially detrimental.
Reference-based metrics can also be used for reranking, for example, by using Minimum Bayes Risk (MBR) decoding \citep{KumarBy04,FGTL22,FeFRSONM22}, which can bias the output toward a particular metric.
While techniques like MBR approximate the reference-based reward function, if a system is evaluated with a reference-free metric, the exact reference-free reward function can be directly optimized during inference.

\section{Conclusion}
In this work, we have argued that reference-free metrics are inherently limited in their ability to evaluate generated text.
Because they are equivalent to generation models, they can be directly optimized at test time, they are biased toward their own models' outputs and outputs from similar models, and they can be biased against higher-quality text, such as human-written references.
Therefore, we recommend against using reference-free metrics as measures of progress on tasks and instead advocate for them to be used as useful statistics to calculate in order to better understand model behavior.

\section*{Acknowledgments}
The authors would like to thank the anonymous ACL ARR reviewers as well as Markus Freitag and Colin Cherry for their insightful feedback on our work.
Their comments raised many interesting points and have helped improved the quality of our final publication.

This work was supported by Contracts FA8750-19-2-1004 and FA8750-19-2-0201 with the US Defense Advanced Research Projects Agency (DARPA). Approved for Public Release, Distribution Unlimited. The views expressed are those of the authors and do not reflect the official policy or position of the Department of Defense or the U.S. Government.

This research is based upon work supported in part by the Oﬃce of the Director of National Intelligence (ODNI), Intelligence Advanced Research Projects Activity (IARPA), via IARPA Contract No. 2019-19051600006 under the BETTER Program. The views and conclusions contained herein are those of the authors and should not be interpreted as necessarily representing the oﬃcial policies, either expressed or implied, of ODNI, IARPA, the Department of Defense, or the U.S. Government. The U.S. Government is authorized to reproduce and distribute reprints for governmental purposes notwithstanding any copyright annotation therein.

This research is supported by a Focused Award from Google.

The second author is supported by the Eric and Wendy Schmidt Postdoctoral Award for Women in Mathematical and Computing Sciences.
\section*{Limitations}

Our work argues that reference-free metrics are theoretically biased toward models which are similar to the metrics' underlying models and biased against human-written output, and we show experimental evidence to support this argument.
While our argument applies to the class of reference-free metrics regardless of the task, our experiments use only one translation and two summarization datasets, so the extent to which this problem can be experimentally demonstrated on other tasks is not known.
However, we suspect it is possible.

Then, although we point out the limitations and biases of reference-free metrics, we do not provide a clear alternative for how to automatically evaluate inherently reference free aspects of text that would not have these problems.
It is not clear if this is possible or how to proceed.

Our analysis in \S\ref{sec:inference_eval} uses reference-based metrics as surrogates for human judgments to determine system output quality.
Ideally we would use humans to determine the true quality of the outputs from our inference algorithms, but that is out of the scope for this paper.

\bibliography{bibliography}
\bibliographystyle{acl_natbib}

\appendix

\section{Implementation Details}
\label{sec:implementation_details}
Our experiments were run on a single Titan RTX GPU with 24 GB of memory.
Since the experiments did not require training models and are limited to using metrics to score model outputs, they were relatively inexpensive;
most experiments completed in several hours.

The ROUGE and BLEU implementations are from SacreROUGE \citep{DeutschRo20} and SacreBLEU \citep{Post18}, respectively.
We also used SacreROUGE for \qaeval{}.
For BLEURT, we use the BLEURT-base-128 version.
BERTScore uses the embeddings from RoBERTa-Large for English and the default contextual embeddings for the non-English languages. \citep{LOGDMJCLLZS19}.
For \questeval{}, we used version v0.1.1 of the corresponding library\footnote{\url{https://github.com/ThomasScialom/QuestEval}} based on the authors' recommendation in their documentation.
The Prism-src and \cometqe{} implementations were based on the authors' original code.
All of the metrics implementations can be found in the Repro library \citep{DeutschRo22c}.
\section{Additional Results}
\label{sec:additional_results}

This Section contains additional results to supplement the main body of the paper.

Table~\ref{tab:inference_table} contains the Figure numbers for additional combinations of reference-free metrics, inference algorithms, and reference-based metrics to what was presented in \S\ref{sec:inference_eval}.
In every setup, the inference procedure generated outputs which performed the best compared to all of the other available systems in the datasets.

Table~\ref{tab:prism_xbleu} contains the correlations calculated between the systems' Prism-src scores and their reference-based scores in which the pseudo-reference is determined by running an inference algorithm on Prism-src.
This experiment was explained in \S\ref{sec:pseudo_ref}.
Tables~\ref{tab:comet_xbleu} and \ref{tab:questeval_xrouge} contain the same results but for \cometqe{} and \questeval{}.

\begin{table*}
    \centering
    \begin{tabular}{llll}
        \toprule
        \bf Reference-Free Metric &
        \bf Optimization Method &
        \bf Reference-Based Metric &
        \bf Figure \\
        \midrule
        Prism-src & Direct Optimization (\S\ref{sec:direct_opt}) & BLEURT & Fig.~\ref{fig:ranking_prism_bleurt_all} \\
        Prism-src & Direct Optimization (\S\ref{sec:direct_opt}) & BLEU & Fig.~\ref{fig:ranking_prism_bleu_all} \\
        Prism-src & Direct Optimization (\S\ref{sec:direct_opt}) & BERTScore & Fig.~\ref{fig:ranking_prism_bertscore_all} \\
        Prism-src & Reranking (\S\ref{sec:reranking}) & BLEURT & Fig.~\ref{fig:ranking_prism_rerank_bleurt} \\
        Prism-src & Reranking (\S\ref{sec:reranking}) & BLEU & Fig.~\ref{fig:ranking_prism_rerank_bleu} \\
        Prism-src & Reranking (\S\ref{sec:reranking}) & BERTScore & Fig.~\ref{fig:ranking_prism_rerank_bertscore} \\
        \cometqe{} & Reranking (\S\ref{sec:reranking}) & BLEURT & Fig.~\ref{fig:ranking_comet_bleurt} \\
        \cometqe{} & Reranking (\S\ref{sec:reranking}) & BLEU & Fig.~\ref{fig:ranking_comet_bleu} \\
        \cometqe{} & Reranking (\S\ref{sec:reranking}) & BERTScore & Fig.~\ref{fig:ranking_comet_bertscore} \\
        QuestEval & Greedy Extractive (\S\ref{sec:greedy_ext}) & ROUGE & Fig.~\ref{fig:ranking_questeval_rouge} \\
        QuestEval & Greedy Extractive (\S\ref{sec:greedy_ext}) & BERTScore & Fig.~\ref{fig:ranking_questeval_bertscore} \\
        QuestEval & Greedy Extractive (\S\ref{sec:greedy_ext}) & QAEval & Fig.~\ref{fig:ranking_questeval_qaeval} \\
        QuestEval & Reranking (\S\ref{sec:reranking}) & ROUGE & Fig.~\ref{fig:ranking_questeval_rerank_rouge} \\
        QuestEval & Reranking (\S\ref{sec:reranking}) & BERTScore & Fig.~\ref{fig:ranking_questeval_rerank_bertscore} \\
        QuestEval & Reranking (\S\ref{sec:reranking}) & QAEval & Fig.~\ref{fig:ranking_questeval_rerank_qaeval} \\
        \bottomrule
    \end{tabular}
    \caption{
        The Figure on each row corresponds to using the optimization method (\S\ref{sec:metric_opt}) to generate an output with the highest reference-free metric score, which is then evaluated with the reference-free and reference-based metric.
        In each Figure, the systems' reference-based scores are on the left y-axis, the reference-free score on the right y-axis, the outputs which were generated by running inference on the reference-free metric in blue, and the reference text's reference-free metric score in the red ``x.''
    }
    \label{tab:inference_table}
\end{table*}

\begin{table*}[t]
    \centering
    \begin{adjustbox}{width=\textwidth}
    \begin{tabular}{lccccccccccccccccc}
        \toprule
        \bf Metric & \bf de-cs & \bf de-en & \bf de-fr & \bf en-cs & \bf en-de & \bf en-fi & \bf en-kk & \bf en-lt & \bf en-ru & \bf en-zh & \bf fi-en & \bf fr-de & \bf kk-en & \bf lt-en & \bf ru-en & \bf zh-en & \bf Avg. \\
        \midrule
        \multicolumn{17}{l}{\emph{Direct Optimization (\S\ref{sec:direct_opt})}} \\
        BLEURT & 0.99 & 0.97 & 0.69 & 0.92 & 0.99 & 0.98 & 0.99 & 0.93 & 0.61 & 0.73 & 0.99 & 0.16 & 0.98 & 0.99 & 0.84 & 0.99 & 0.86 \\
        BLEU & 0.95 & 0.92 & 0.62 & 0.94 & 0.96 & 0.93 & 0.96 & 0.91 & 0.77 & 0.55 & 0.98 & 0.47 & 0.86 & 0.97 & 0.82 & 0.98 & 0.85 \\
        BERTScore & 0.98 & 1.00 & 0.90 & 1.00 & 0.99 & 1.00 & 0.99 & 0.99 & 0.80 & 0.87 & 0.99 & 0.82 & 1.00 & 0.97 & 0.95 & 0.99 & 0.95 \\
        \midrule
        \multicolumn{17}{l}{\emph{Reranking (\S\ref{sec:reranking})}} \\
        BLEURT & - & 0.92 & - & - & 0.94 & - & - & - & 0.66 & - & - & - & - & - & 0.85 & - & 0.84 \\
        BLEU & - & 0.83 & - & - & 0.89 & - & - & - & 0.82 & - & - & - & - & - & 0.80 & - & 0.83 \\
        BERTScore & - & 0.95 & - & - & 0.97 & - & - & - & 0.83 & - & - & - & - & - & 0.95 & - & 0.92 \\
        \bottomrule
    \end{tabular}
    \end{adjustbox}
    \caption{The correlation between (1) the Prism-src score for the systems submitted to WMT'19 and (2) the same systems' reference-based metric (left column) score calculated against the pseudo-reference for Prism-src that was generated using the direct optimization inference algorithm (top) or reranking (bottom).}
    \label{tab:prism_xbleu}
\end{table*}
\begin{table}[t]
    \centering
    \begin{adjustbox}{width=\columnwidth}
    \begin{tabular}{lccccc}
        \toprule
        \bf Metric & \bf de-en & \bf en-de & \bf en-ru & \bf ru-en & \bf Avg. \\
        \midrule
        BLEURT & 0.98 & 0.96 & 0.61 & 0.76 & 0.83 \\
        BLEU & 0.91 & 0.95 & 0.86 & 0.80 & 0.88 \\
        BERTScore & 0.99 & 0.97 & 0.84 & 0.87 & 0.92 \\
        \bottomrule
    \end{tabular}
    \end{adjustbox}
    \caption{The correlation between (1) the \cometqe{} score for the systems submitted to WMT'19 and (2) the same systems' reference-based metric (left column) score calculated using the pseudo-reference for \cometqe{} that was generated using the reranking inference algorithm.}
    \label{tab:comet_xbleu}
\end{table}
\begin{table}[t]
    \centering
    \begin{adjustbox}{width=\columnwidth}
    \begin{tabular}{lccc}
        \toprule
        \bf Metric & \bf \summeval{} & \bf \realsumm{} & \bf Avg. \\
        \midrule
        \multicolumn{4}{l}{\emph{Greedy Extractive (\S\ref{sec:greedy_ext})}} \\
        ROUGE-2 & 0.85 & 0.83 & 0.84 \\
        BERTScore & 0.62 & 0.85 & 0.74 \\
        QAEval & 0.86 & 0.89 & 0.88 \\
        \midrule
        \multicolumn{4}{l}{\emph{Reranking (\S\ref{sec:reranking})}} \\
        ROUGE-2 & 0.77 & 0.77 & 0.77 \\
        BERTScore & 0.19 & 0.78 & 0.49 \\
        QAEval & 0.87 & 0.85 & 0.86 \\
        \bottomrule
    \end{tabular}
    \end{adjustbox}
    \caption{The correlation between (1) the \questeval{} score for the systems included in the \realsumm{} and \summeval{} datasets and (2) the same systems' reference-based metric (left column) score calculated against the pseudo-reference for \questeval{} that was generated using the greedy extractive summarization inference algorithm (top) or reranking (bottom).}
    \label{tab:questeval_xrouge}
\end{table}

\begin{figure*}[t]
    \centering
    \includegraphics[width=\textwidth]{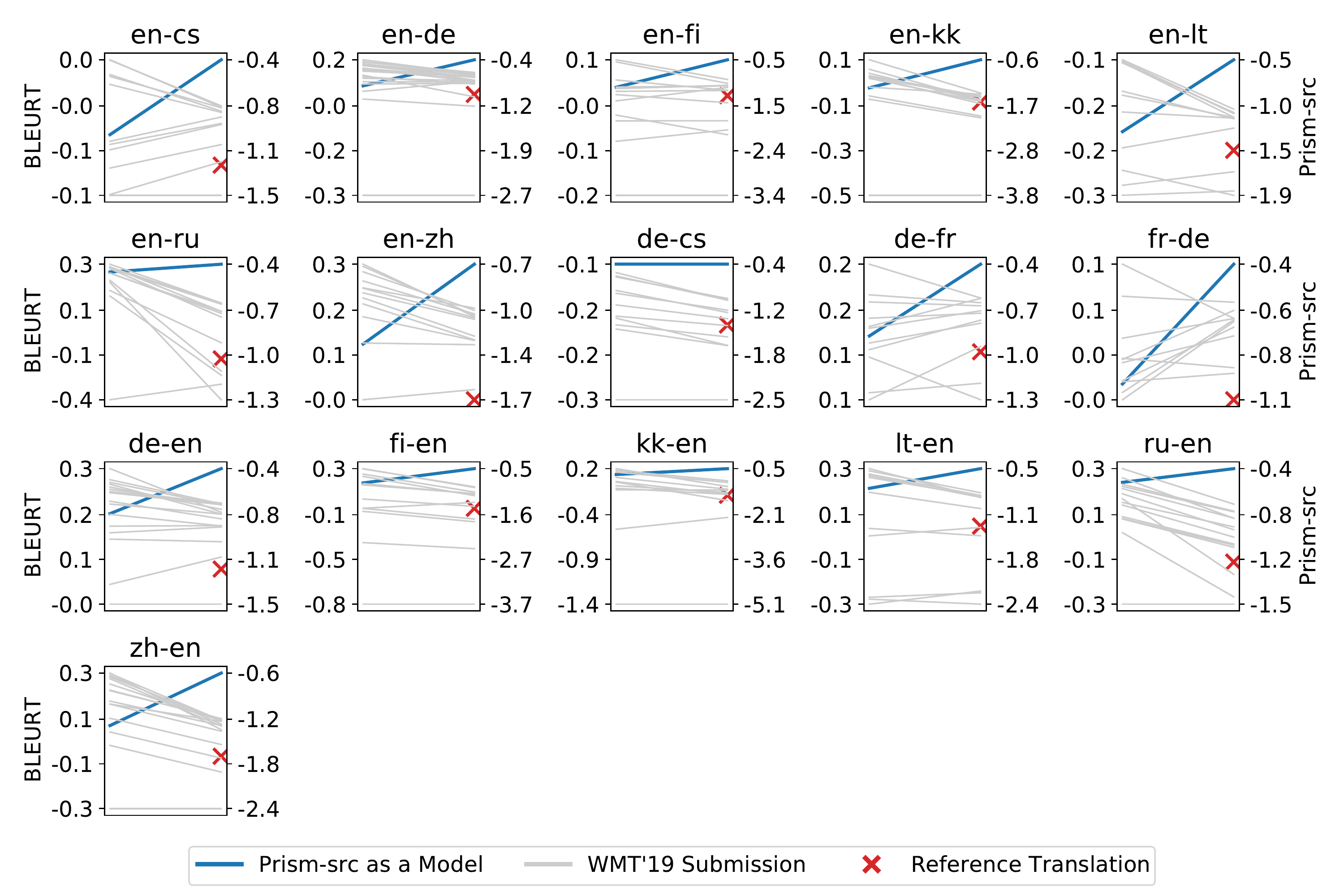}
    \caption{See Table~\ref{tab:inference_table} for a description of this Figure.}
    \label{fig:ranking_prism_bleurt_all}
\end{figure*}
\begin{figure*}[t]
    \centering
    \includegraphics[width=\textwidth]{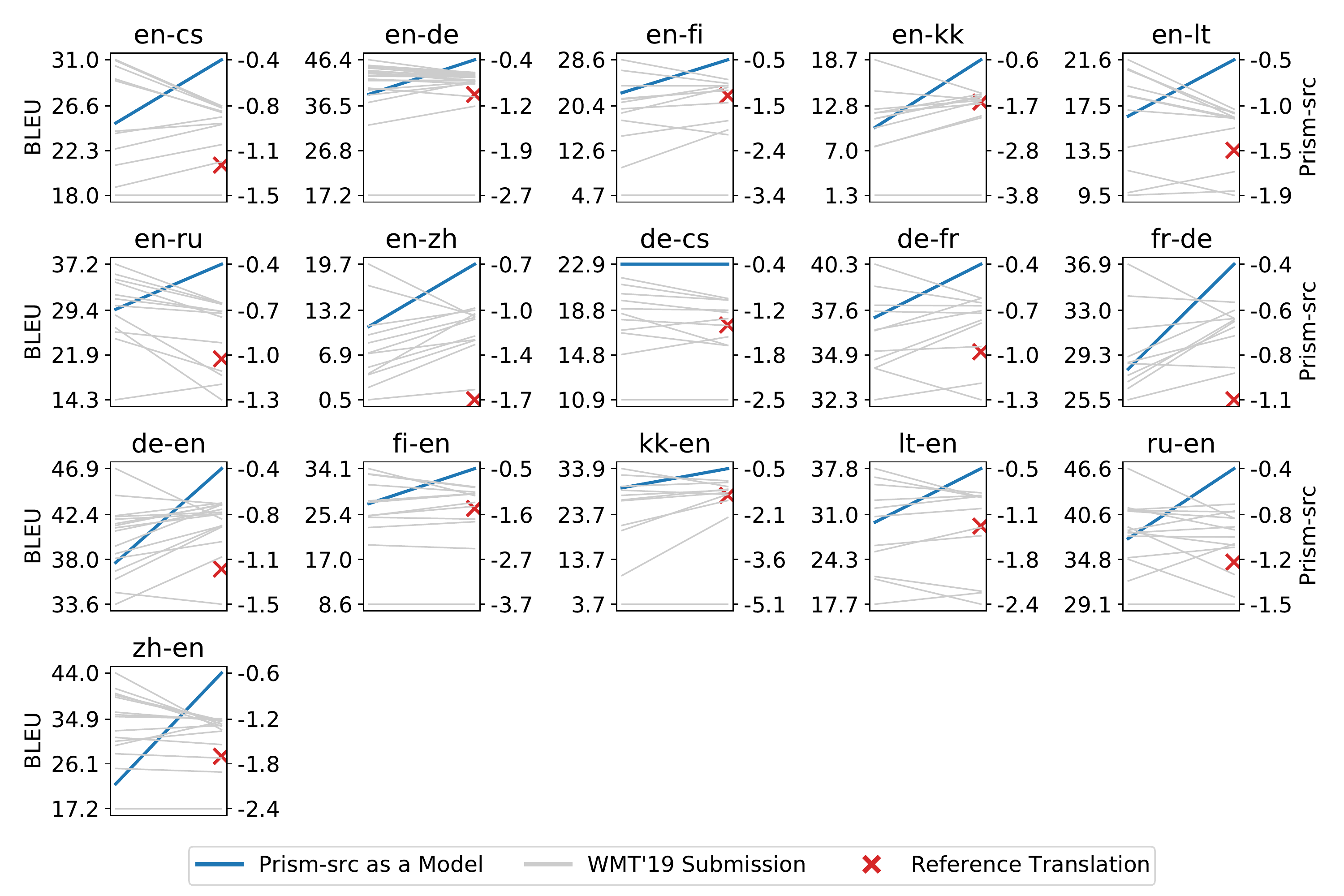}
    \caption{
    See Table~\ref{tab:inference_table} for a description of this Figure.
    }
    \label{fig:ranking_prism_bleu_all}
\end{figure*}
\begin{figure*}[t]
    \centering
    \includegraphics[width=\textwidth]{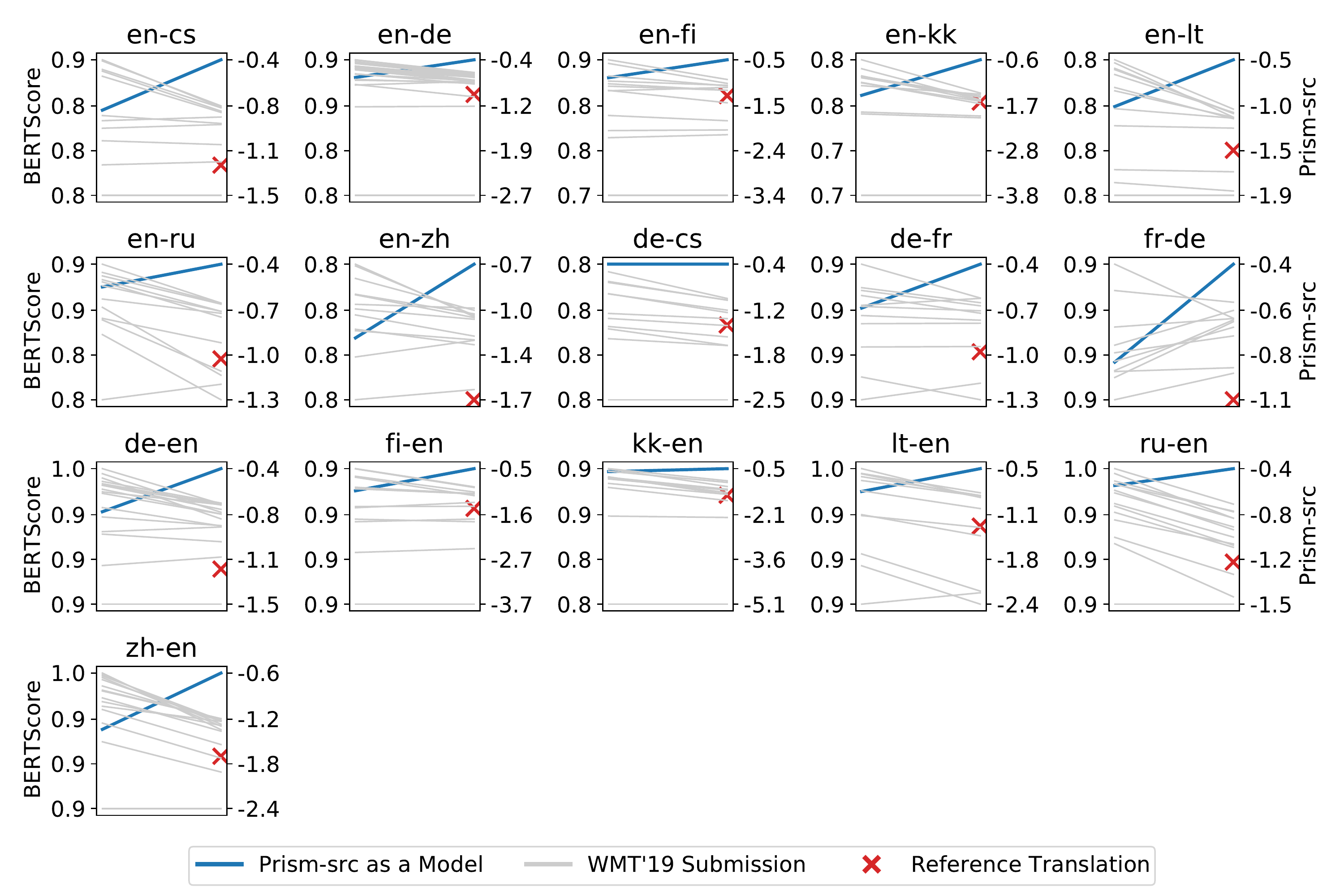}
    \caption{See Table~\ref{tab:inference_table} for a description of this Figure.
    }
    \label{fig:ranking_prism_bertscore_all}
\end{figure*}

\begin{figure}[t]
    \centering
    \includegraphics[width=\columnwidth]{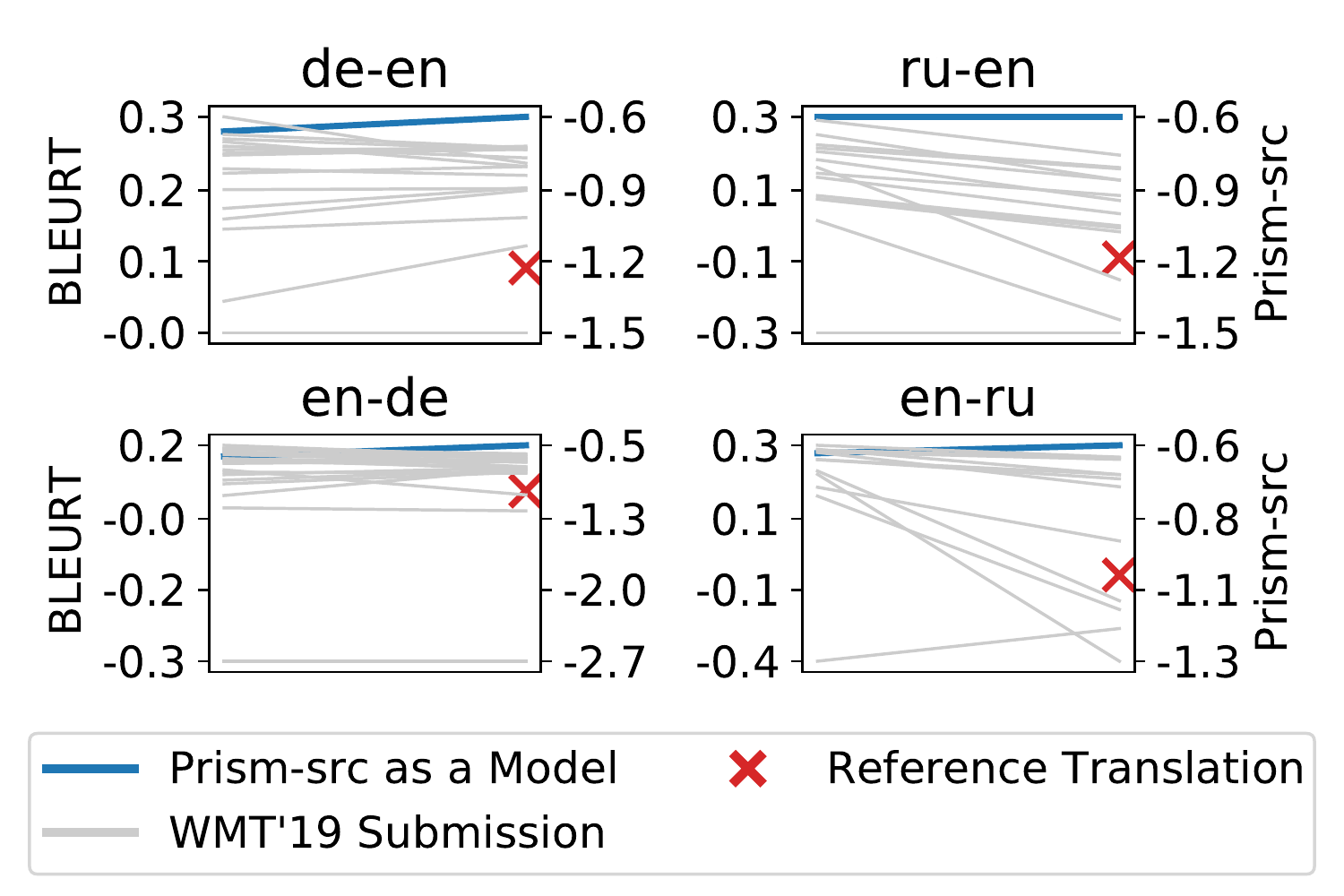}
    \caption{See Table~\ref{tab:inference_table} for a description of this Figure.}
    \label{fig:ranking_prism_rerank_bleurt}
\end{figure}
\begin{figure}[t]
    \centering
    \includegraphics[width=\columnwidth]{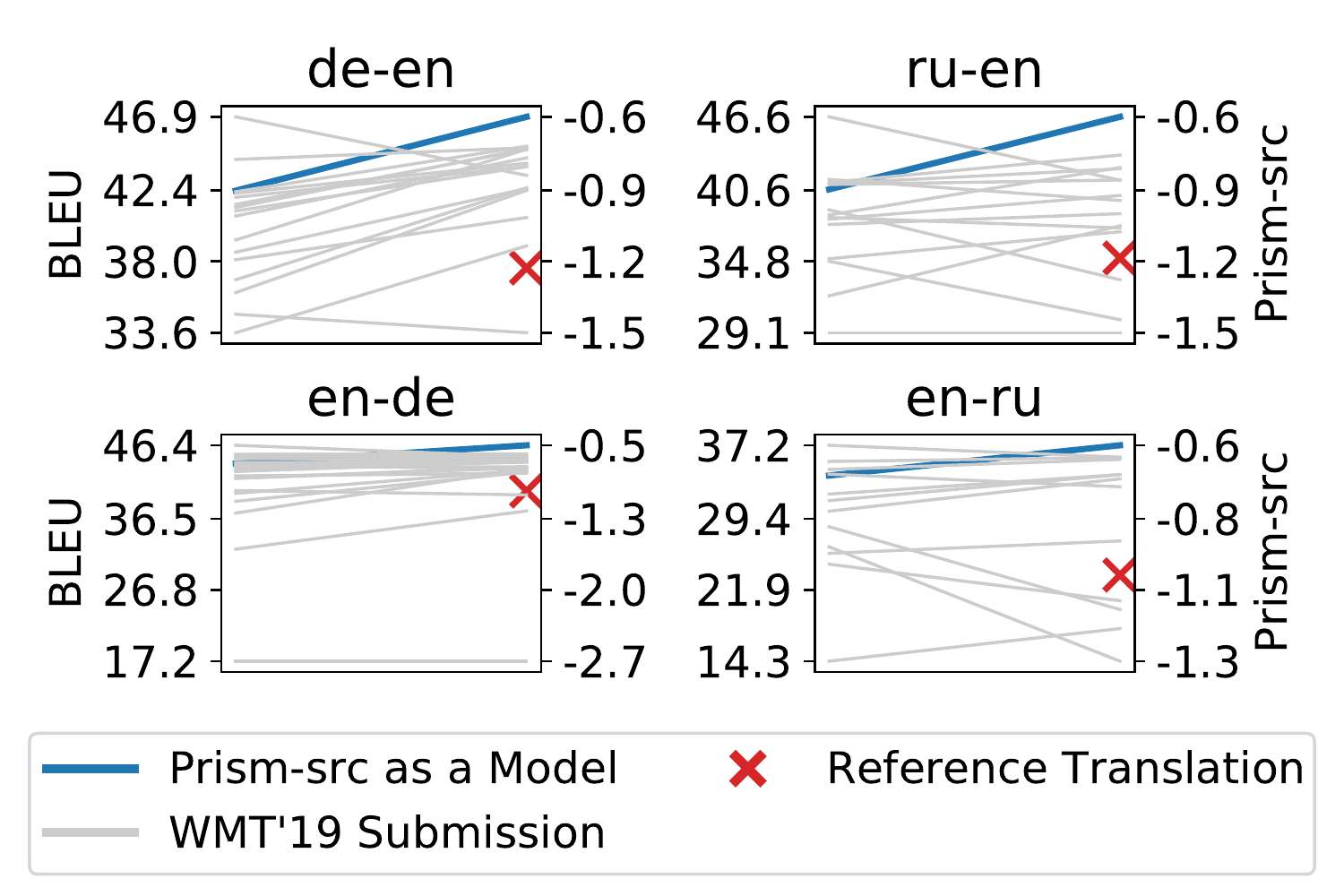}
    \caption{See Table~\ref{tab:inference_table} for a description of this Figure.}
    \label{fig:ranking_prism_rerank_bleu}
\end{figure}
\begin{figure}[t]
    \centering
    \includegraphics[width=\columnwidth]{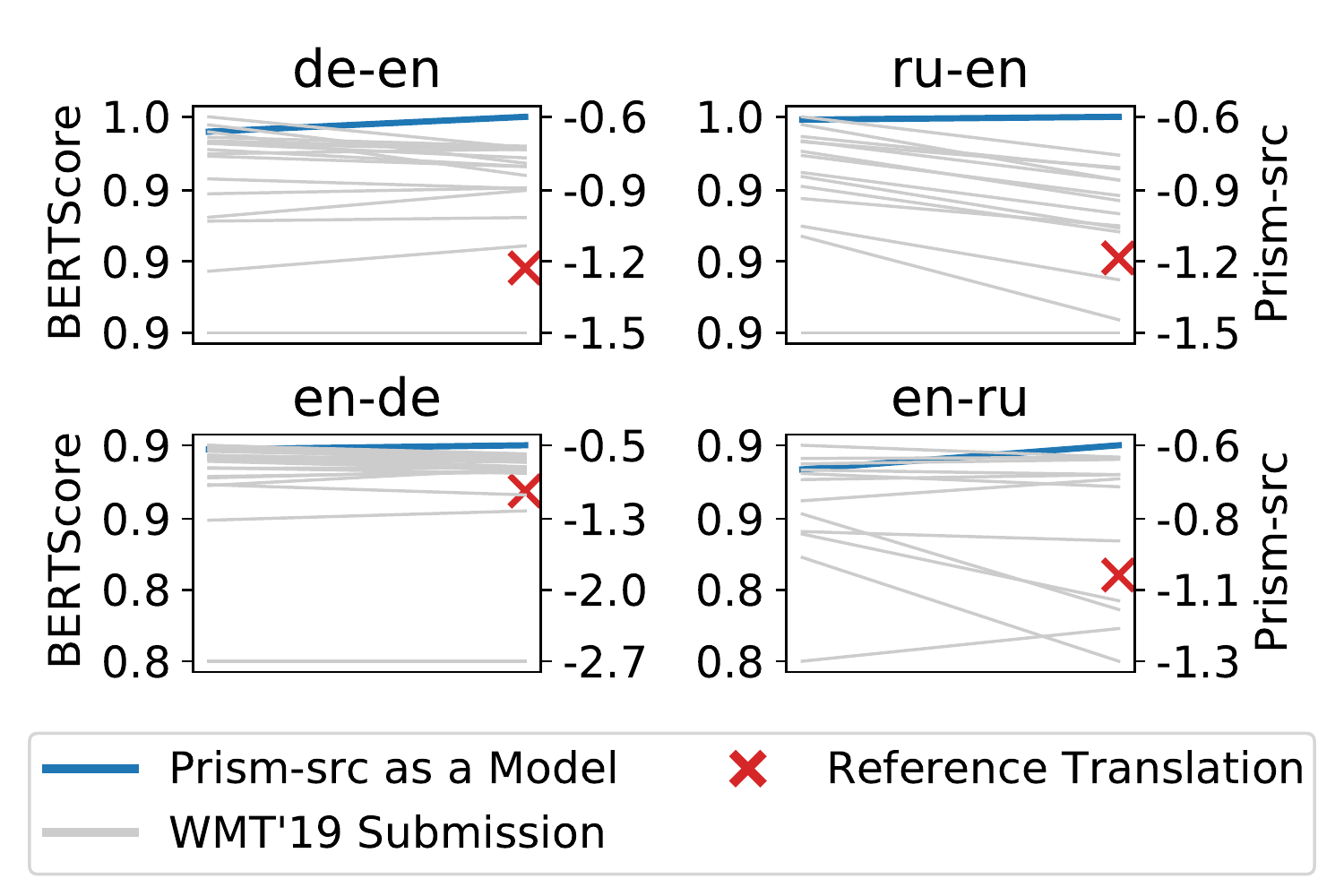}
    \caption{See Table~\ref{tab:inference_table} for a description of this Figure.}
    \label{fig:ranking_prism_rerank_bertscore}
\end{figure}

\begin{figure}[t]
    \centering
    \includegraphics[width=\columnwidth]{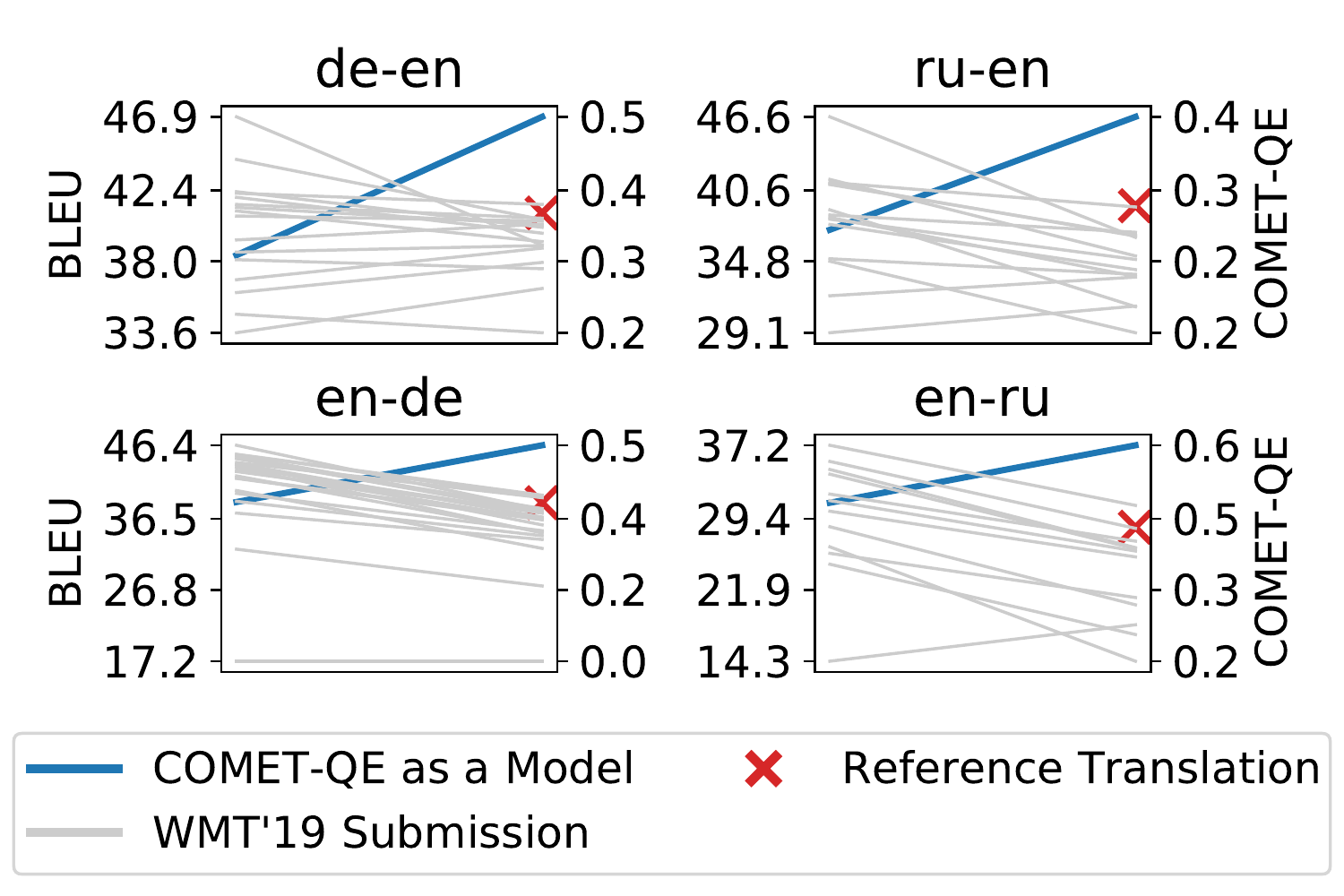}
    \caption{See Table~\ref{tab:inference_table} for a description of this Figure.}
    \label{fig:ranking_comet_bleu}
\end{figure}
\begin{figure}[t]
    \centering
    \includegraphics[width=\columnwidth]{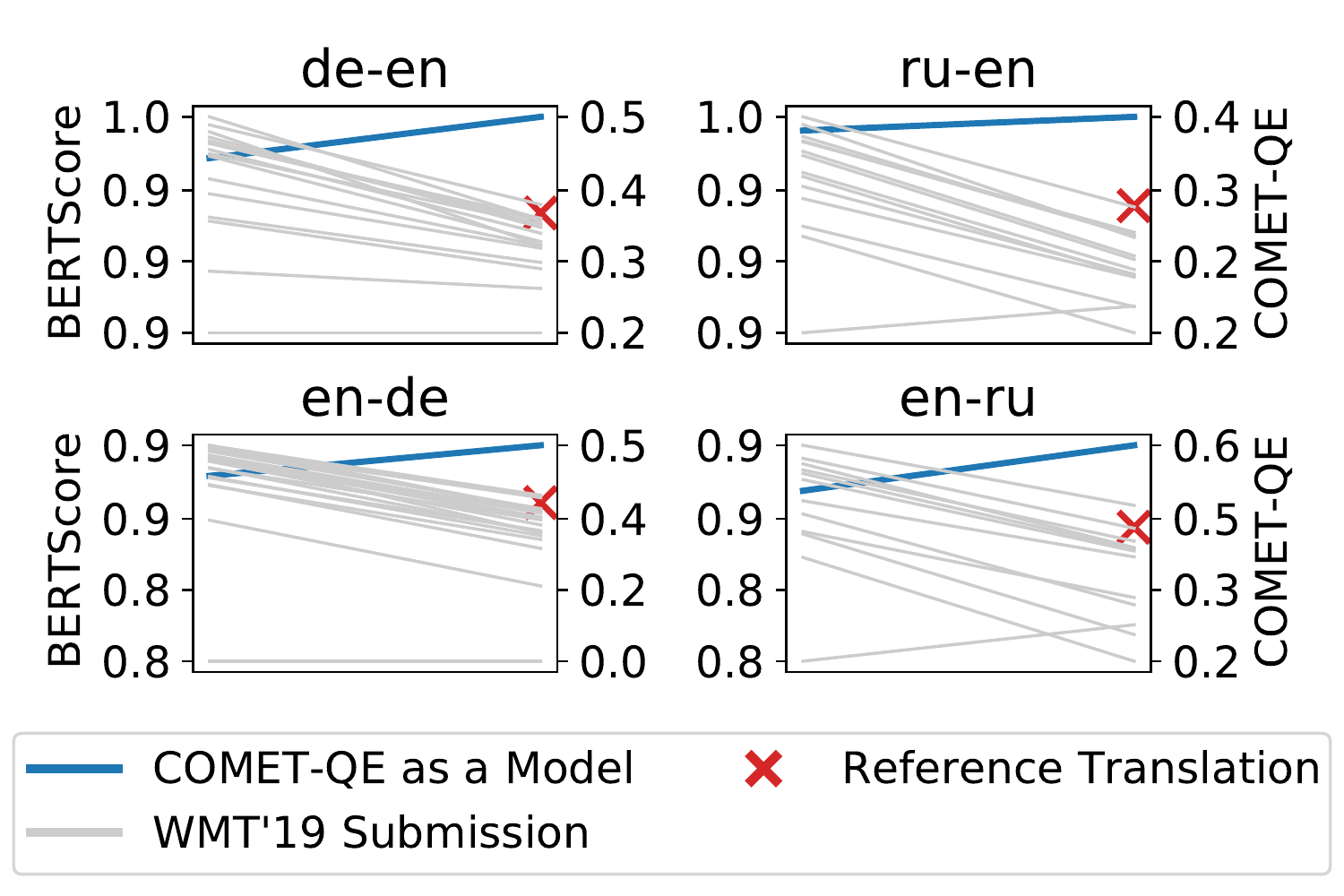}
    \caption{See Table~\ref{tab:inference_table} for a description of this Figure.}
    \label{fig:ranking_comet_bertscore}
\end{figure}

\begin{figure}[t]
    \centering
    \includegraphics[width=\columnwidth]{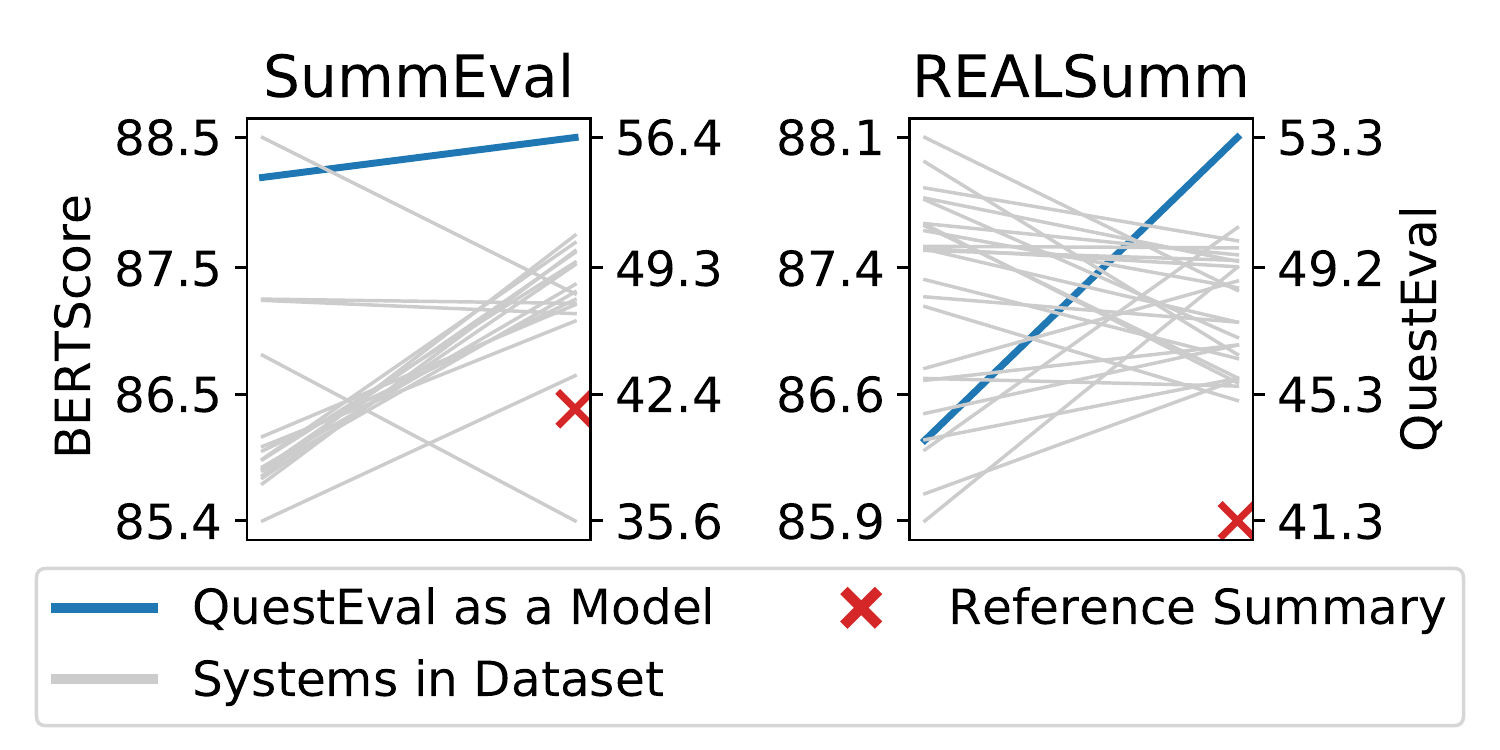}
    \caption{
        See Table~\ref{tab:inference_table} for a description of this Figure.
    }
    \label{fig:ranking_questeval_bertscore}
\end{figure}
\begin{figure}[t]
    \centering
    \includegraphics[width=\columnwidth]{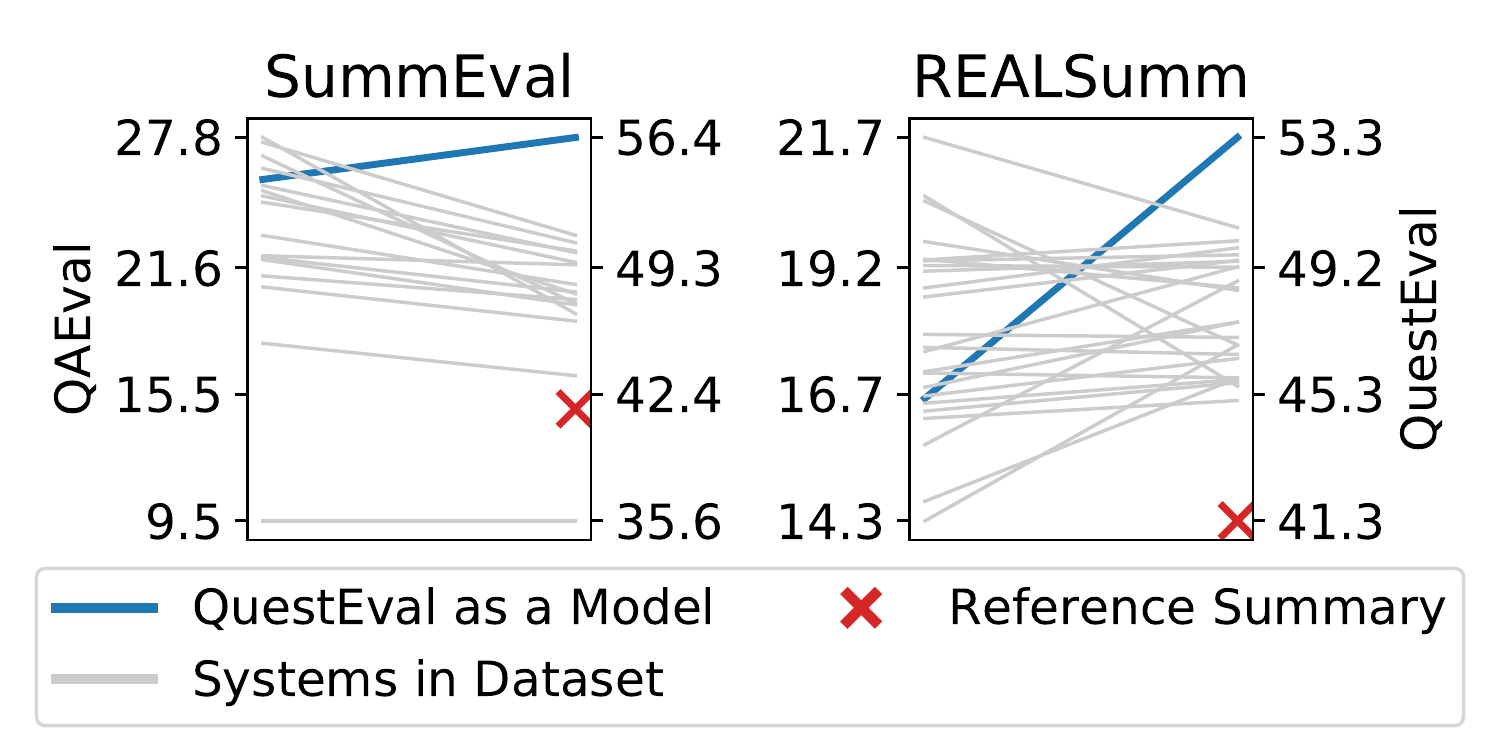}
    \caption{
        See Table~\ref{tab:inference_table} for a description of this Figure.
    }
    \label{fig:ranking_questeval_qaeval}
\end{figure}

\begin{figure}[t]
    \centering
    \includegraphics[width=\columnwidth]{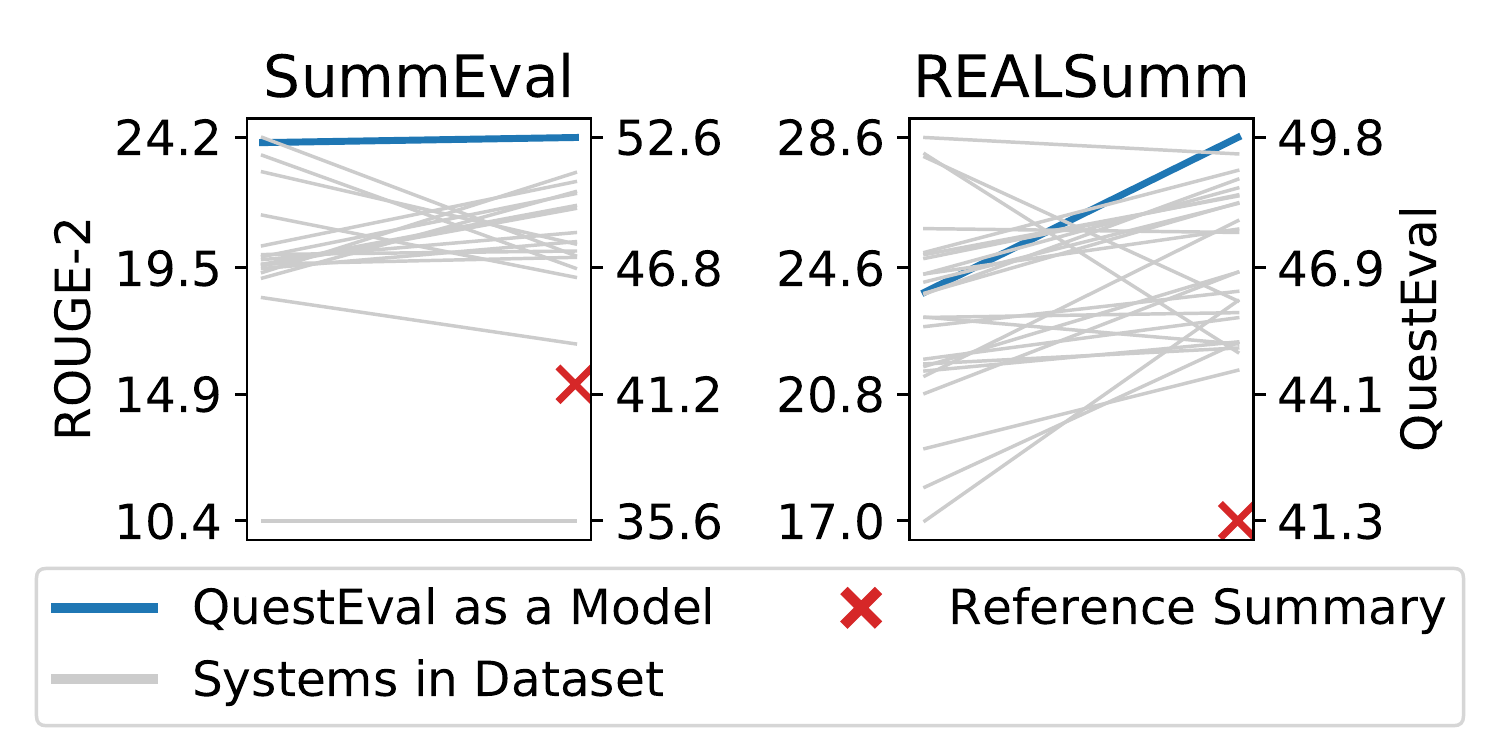}
    \caption{
        See Table~\ref{tab:inference_table} for a description of this Figure.
    }
    \label{fig:ranking_questeval_rerank_rouge}
\end{figure}
\begin{figure}[t]
    \centering
    \includegraphics[width=\columnwidth]{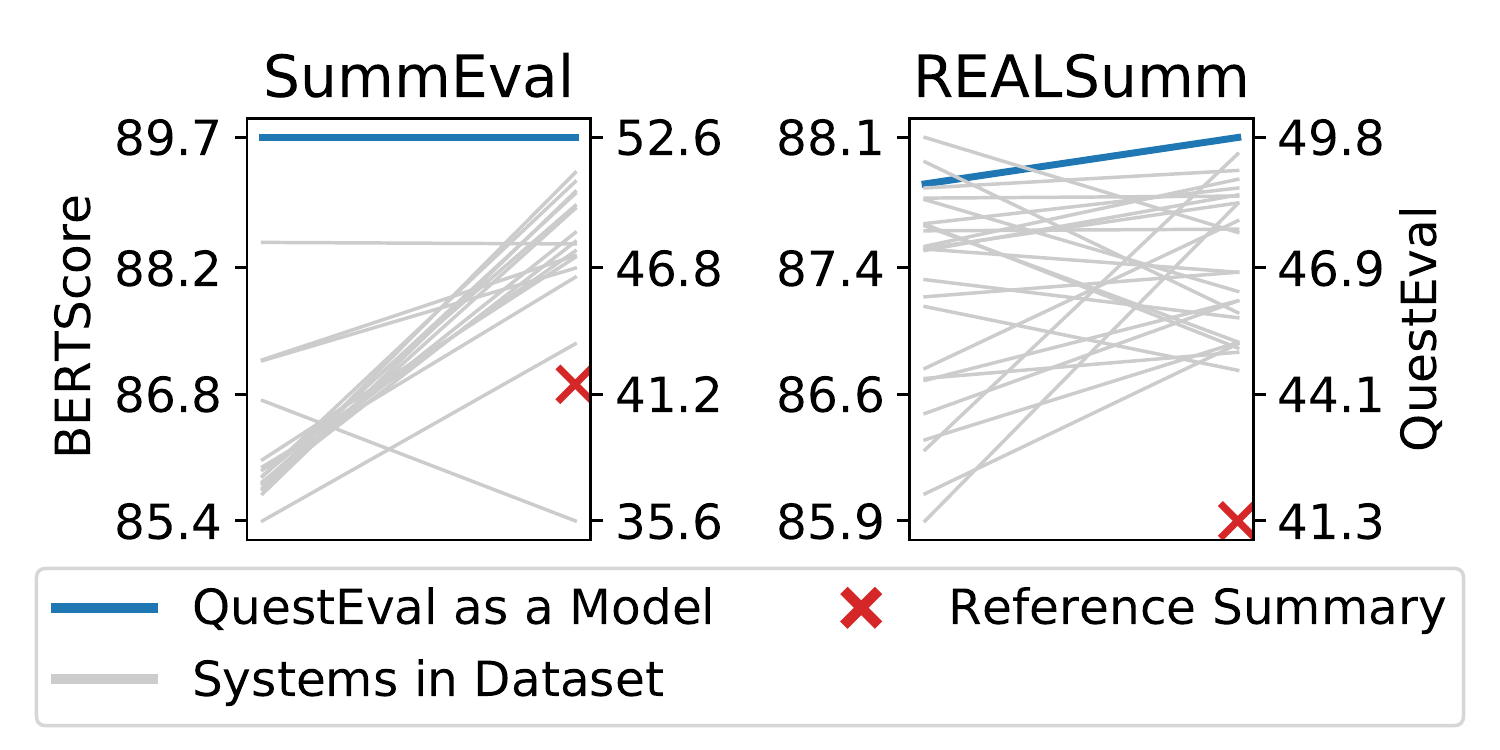}
    \caption{
        See Table~\ref{tab:inference_table} for a description of this Figure.
    }
    \label{fig:ranking_questeval_rerank_bertscore}
\end{figure}
\begin{figure}[t]
    \centering
    \includegraphics[width=\columnwidth]{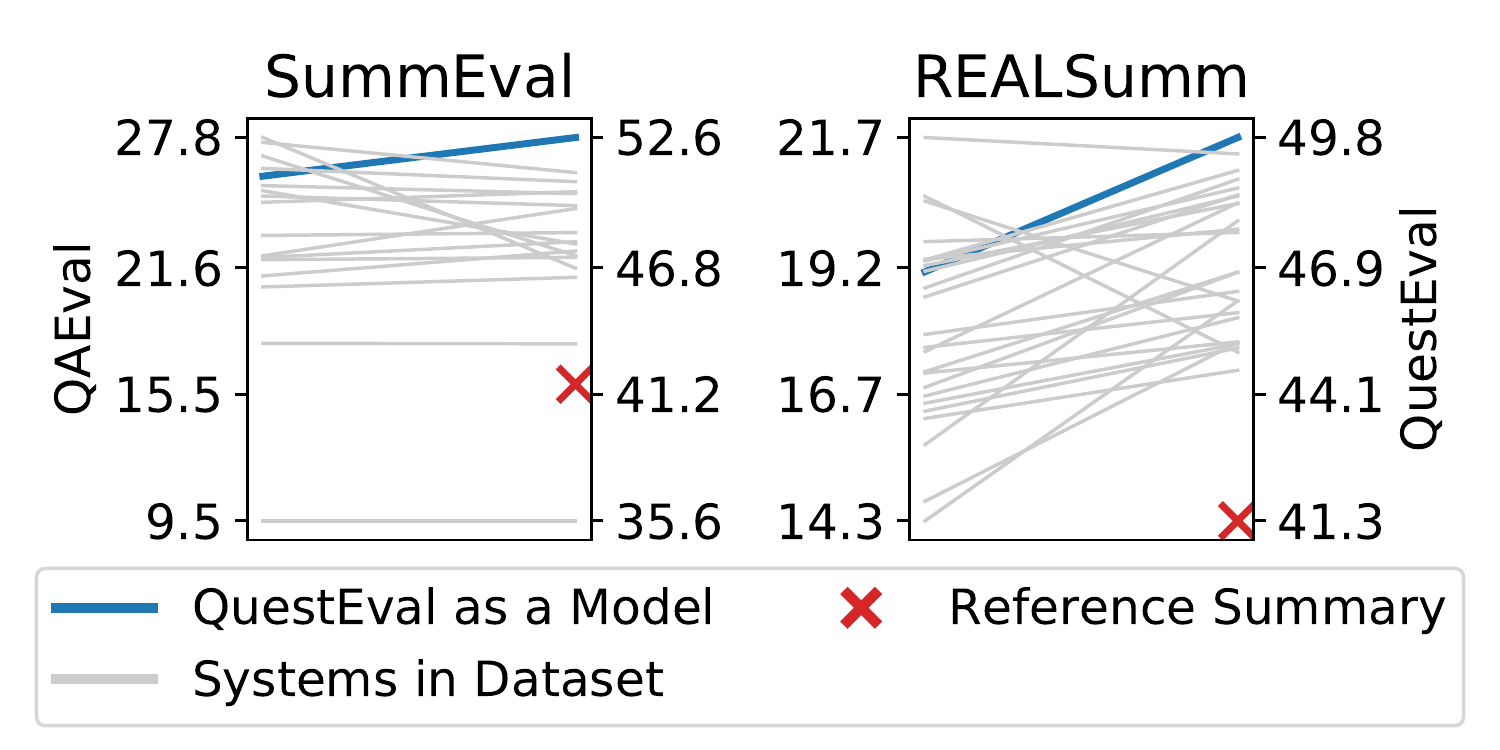}
    \caption{
        See Table~\ref{tab:inference_table} for a description of this Figure.
    }
    \label{fig:ranking_questeval_rerank_qaeval}
\end{figure}

\end{document}